\documentclass[conference]{IEEEtran}
\IEEEoverridecommandlockouts
\usepackage{cite}
\usepackage{amsmath,amssymb,amsfonts}
\usepackage{dsfont}
\usepackage{array}
\usepackage{algorithmic}
\usepackage{graphicx}
\usepackage{textcomp}
\usepackage{xcolor}
\usepackage[caption = false]{subfig}
\usepackage{soul}
\usepackage{bbding}
\usepackage{booktabs}
\usepackage{tabularx}
\def\BibTeX{{\rm B\kern-.05em{\sc i\kern-.025em b}\kern-.08em
    T\kern-.1667em\lower.7ex\hbox{E}\kern-.125emX}}
\begin{document}

% ------------------------------------------------------------------
\title{A Contrario multi-scale anomaly detection method for industrial quality inspection

\thanks{This work was partially funded by a graduate scholarship from Agencia Nacional de Investigación e Innovación, Uruguay.}
}

% ------------------------------------------------------------------
\author{\IEEEauthorblockN{Matías Tailanian}
\IEEEauthorblockA{\textit{Digital Sense \&} \\
\textit{Universidad de la República}\\
Montevideo, Uruguay \\
mtailanian@digitalsense.ai}
\and
\IEEEauthorblockN{Pablo Musé}
\IEEEauthorblockA{\textit{IIE, Facultad de Ingenier\'ia} \\
\textit{Universidad de la República}\\
Montevideo, Uruguay \\
pmuse@fing.edu.uy }
\and
\IEEEauthorblockN{Álvaro Pardo}
\IEEEauthorblockA{\textit{Universidad Cat\'olica del Uruguay \&} \\
\textit{Digital Sense}\\
Montevideo, Uruguay \\
apardo@ucu.edu.uy}
}

\maketitle
% ------------------------------------------------------------------
\begin{abstract}
% ------------------------------------------------------------------
Anomalies can be defined as any non-random structure which deviates from normality. Anomaly detection methods reported in the literature are numerous and diverse, as  what is considered anomalous usually varies depending on particular scenarios and applications. In this work we propose an \textit{a contrario} framework to detect anomalies in images applying statistical analysis to feature maps obtained via convolutions. We evaluate filters learned from the image under analysis via patch PCA, Gabor filters and the feature maps obtained from a pre-trained deep neural network (Resnet). The proposed method is multi-scale and fully unsupervised and is able to detect anomalies in a wide variety of scenarios. While the end goal of this work is the detection of subtle defects in leather samples for the automotive industry, we show that the same algorithm achieves state-of-the-art results in public anomalies datasets.
\end{abstract}


% ------------------------------------------------------------------
\section{Introduction}
% ------------------------------------------------------------------

Anomaly detection is an active field that has been studied for decades, motivated by a wide variety of practical applications ranging from robotics and security to health care. One of the most relevant applications is automatic or aided product quality inspection in industrial production. 

In most situations it is very difficult or even impossible to collect a statistically relevant sample of all types of anomalies, given its rare nature and intra-class variability. In addition, in most cases anomalies are very subtle, and the line that separates them from normality is thin and fuzzy. These kinds of problems are typically formulated as one-class classification problems~\cite{oneclass}, where the strategy is based on learning and characterizing the statistics of normality. The key advantage of this scheme is that only ``normal'' (i.e. anomaly free) samples are needed for the training process. Then, new tested samples are assigned an anomaly score that represents the deviation of the tested sample to the characterized normality.

\captionsetup{position=top}
\begin{figure}[tb]

\subfloat[]{\includegraphics[width=0.192\linewidth]{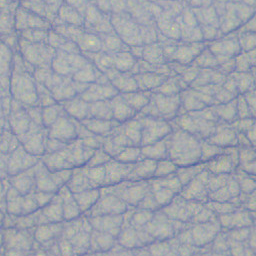}} \hfill
\subfloat[]{\includegraphics[width=0.192\linewidth]{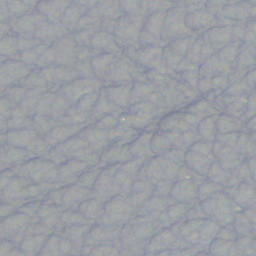}} \hfill
\subfloat[]{\includegraphics[width=0.192\linewidth]{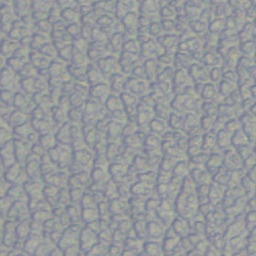}} \hfill
\subfloat[]{\includegraphics[width=0.192\linewidth]{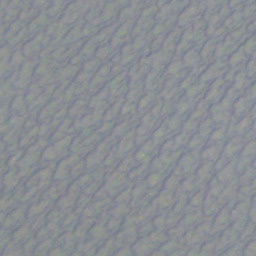}} \hfill
\subfloat[]{\includegraphics[width=0.192\linewidth]{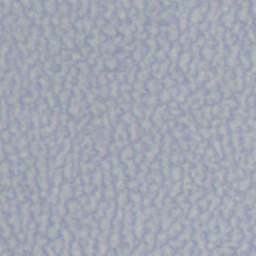}} \hfill \\[-17px]

\captionsetup{position=bottom}
\subfloat[]{\includegraphics[width=0.192\linewidth]{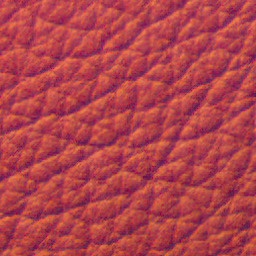}} \hfill
\subfloat[]{\includegraphics[width=0.192\linewidth]{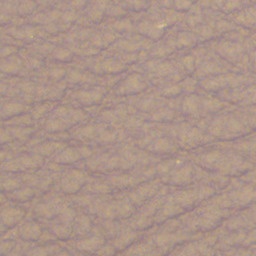}} \hfill
\subfloat[]{\includegraphics[width=0.192\linewidth]{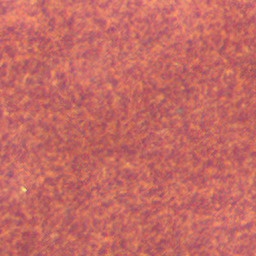}} \hfill
\subfloat[]{\includegraphics[width=0.192\linewidth]{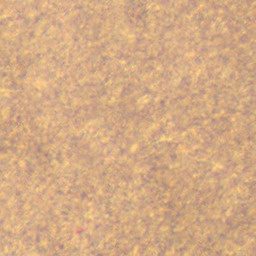}} \hfill
\subfloat[]{\includegraphics[width=0.192\linewidth]{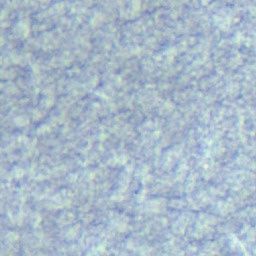}} \hfill \\%[-10px]
    \caption{[Best viewed in color] Motivating application: defect detection in leather samples. Example images showing the diversity of samples. Image (a) and (b) correspond to the same texture, with different engraving strength. The defect present in (b) may not be considered an anomaly if it was in (a). Images (c), (d) and (e) are examples of different textures. (f) and (g) show different color possibilities, and (h), (i) and (j) correspond to the down side of leather samples.}
    % \vspace{-18pt}
    \label{fig:example-images}
\end{figure}

A very different situation arises when normality and abnormality are not absolute notions, but relative to the sample itself. Indeed, in many scenarios, a very same pattern can be considered normal or abnormal when it occurs within two different samples. This situation is depicted in Figure~\ref{fig:example-images}, which shows some examples of the data considered for the application that motivates the present work. In this application, which will be described in detail in Section~\ref{sec:bader}, the goal is to segment defects on processed leather pieces for the automotive industry. The leather could be coloured with a wide variety of tones, may present different textures, and the strength of the texture engraving may also differ. Note for example that the anomalous pattern in Figure~\ref{fig:example-images}(b) should be considered normal if it was present in the sample of Figure~\ref{fig:example-images}(a). 

In summary, the problem we aim to tackle is such that normality can only be modeled from a single image sample, and where we have no prior knowledge whether this sample contains anomalies or not. These conditions frame our solution to a small set of methods, discarding all techniques based on learning from any set of images, and those that extract statistics from a set of anomaly-free samples. To this end, we developed a completely unsupervised algorithm that can handle all this variations and detect anomalies in both sides of the leather.

A well established methodology to deal with unsupervised anomaly detection under these conditions is the \textit{a contrario} approach~\cite{lowe2012perceptual,desolneux2007gestalt}. This methodology is commonly used in anomaly detection and has proven to produce impressive results in many tasks, such as clustering, edges and line segments detection~\cite{cao2008theory, von2008lsd}%,von2008straight}
, general point alignments~\cite{lezama2014contrario}, among others. 
In this work we propose to apply this methodology to detect anomalies, after a step of feature extraction devised to enhance the difference between normal and anomalous patches in the image (Section~\ref{sec:method}).

% {\setlength{\tabcolsep}{1em} {\renewcommand{\arraystretch}{1.}%
% \begin{table}[b]
% \vspace{-15pt}
% \caption{}\label{}
% \vspace{-5pt}
% \centering
\begin{table*}[t]
\caption{\vspace{2pt}Related work comparison showing some of the key elements of each method: how they generate the final anomaly map, whether if they use some pre-trained network or not, how is the usage of normal and abnormal images, the kind of supervision, if they solve a proxy task and the way to achieve the multi-scale analysis.}
\label{tab:related-work}
% \vspace{-10pt}
\scriptsize
\begin{center}
    \begin{tabular}{m{7em}|m{17em}|m{3em}|m{5em}|m{3.3em}|m{2em}|m{4.5em}|m{5.5em}|m{9.4em}}
        
        {\bfseries Method} & {\bfseries Anomaly map} & {\bfseries Pre-trained} & {\bfseries Normal images usage} & {\bfseries Anomaly img use}  & {\bfseries Thr} & {\bfseries Supervision} & {\bfseries Proxy task} & {\bfseries Multi-Scale}\\[5pt]
        \hline  
        \vspace{2pt}
        
\textbf{TextureInsp}~\cite{text} & Negative log-likelihood for trained GMM & No & Fit GMM & No & No & Sup. & No & 4-layer pyramid \\ 
\textbf{SPADE}~\cite{spade} & Euclidean distance to normal samples & Resnet & kNN & No & No & Sup. & No & DNN scales \\
\textbf{DictSimilarity}~\cite{dict} & Distance to normality & Yes & Build dict. & No & No & Sup. & No & Different patch sizes \\ 
\textbf{FCDD}~\cite{fcdd} & Activation layer from network & No & FCN & Yes & No & Sup. & No & DNN scales \\ 
\textbf{PaDiM}~\cite{padim} & Mahalanobis distance by patch & Resnet & Mah. dist. & No & No & Self-Sup. & No & DNN scales \\
\textbf{PatchSVDD}~\cite{patch-svdd} & Distance to normal samples & No & Encoder & No & No & Self-Sup. & Patch-gather & Arch design \\ 
\textbf{RotDet}~\cite{rotnet} & KDE at each patch location & No & DNN & No & No & Self-Sup. & Rotation & DNN scales \\ 
\textbf{STPM}~\cite{student-teacher} & Student-teacher L2 distance & Resnet & Student net & No & No & Self-Sup. & No & DNN scales \\ 
\textbf{DetInNoise}~\cite{ehret2019reduce} & Log-NFA map & Optional & No & No & Yes & UnSup. & No & Downsampled images \\[2pt]
\hline
\vspace{2pt}
\textbf{SSIM-AE}~\cite{bergmann-ssim} & Reconstruction error & No & AE & No & No & Self-Sup. & No & AE architecture \\ 
\textbf{AnoGan}~\cite{anogan} & Reconstruction error & No & GAN & No & No & Self-Sup. & No & GAN architecture \\ 
\textbf{InTra}~\cite{intra} & Reconstruction error & No & Transformer & No & No & Self-Sup. & Inpainting & Downsampled images \\ 
\textbf{CutPaste}~\cite{cutpaste} & One class over representation & No & DNN & No & No & Self-Sup. & Cut-Paste & Replicating \\ 
\textbf{DFR}~\cite{dfr} & Reconstruction error & VGG19 & VAE & No & No & Self-Sup. & No & DNN scales \\[2pt]
%\textbf{IGD}~\cite{igd} & Local and global reconstruction error & Optional & AE & No & No & Self-Sup. & No & Whole image + patches \\ 
% @article{igd,
%   title={Unsupervised Anomaly Detection with Multi-scale Interpolated Gaussian Descriptors},
%   author={Chen, Yuanhong and Tian, Yu and Pang, Guansong and Carneiro, Gustavo},
%   journal={arXiv preprint arXiv:2101.10043},
%   year={2021}
% }
\hline 
\vspace{2pt}
\textbf{GBVS}~\cite{gbvs-harel2007graph} & Equilibrium distribution of markov chain & No & No & No & No & UnSup. & No & Downsampled images \\ 
\textbf{SALICON}~\cite{salicon-huang2015salicon} & Activation layer from network & No & No & Yes & No & Sup. & No & Arch design \\ 
\textbf{DRFI}~\cite{drfi-jiang2013salient} & RF Regression & No & No & Yes & No & Sup. & No & Multi-Level segmentat. \\[2pt]
\hline 
\vspace{2pt}
\textbf{Ours} & Log-NFA map & Optional & No & No & Yes & UnSup. & No & Downsampled images \\ 

    \end{tabular}
\end{center}
\normalsize
% \vspace{-15pt}
\end{table*}
% \end{table}
% }}

Based on the fact that learning a statistical model of anomalies from data 
is not possible due to their very limited number of occurrences, the {\em a contrario} methodology, as other approaches, focuses on the design of a {\em background model} or null hypothesis that characterizes normality. Following the non-accidentalness principle~\cite{witkin1983role}, the anomalies are detected as events such that the probability of occurrence under the background model is so small that they are much more likely to result from another cause. This principle is very intuitive. Suppose we observe a certain specific arrangement or structure in an image. The more rare this arrangement is, the less probable that it corresponds to a mere random arrangement, indicating a possible presence of an anomaly. Therefore, what we need to evaluate is the probability of that arrangement being generated by the background model. Another characteristic that makes the {\em a contrario} framework different and particularly useful, is that it automatically fixes detection thresholds that control the number of false alarms (NFA)~\cite{desolneux2007gestalt,ehret2019reduce,le2019efficient,von2021ground}, allowing not only to detect rare events in very diverse backgrounds but also to associate a rareness score. This score has a clear statistical meaning: it is an estimate of the number of occurrences of an observed event if it was produced by the background model. Thus, obtaining a very low NFA value, means that is very unlikely that the event was generated by the background model, and therefore it is probably a meaningful event in the anomaly detection scheme. \\

In short, in this work we develop a fully unsupervised multi-scale \textit{a contrario} anomaly detection method that proves to be highly versatile with very good results. Our algorithm does not need any normal or anomalous datasets, nor any priors. In practice, this method is parameterless and benefits from all the computational power of neural networks libraries.\\

The remainder of this paper is organized as follows: in Section~\ref{sec:relatedwork} we present an exhaustive review of state-of-the-art methods and compare their key elements. Then, we present our work as a general algorithm to detect anomalies, explaining the method and showing results in Sections~\ref{sec:method}, and~\ref{sec:results}. %, and compare our to state of the art methods in Section~\ref{sec:results}. 
Finally in Section~\ref{sec:bader} we apply the proposed method to our particular industrial application, and we present its results.%, namely the detection of defects in leather for the automotive industry, and show the results obtained.
\vspace{10pt}
% ------------------------------------------------------------------
\section{Related Work}
\label{sec:relatedwork}
% ------------------------------------------------------------------

There is a wide literature on anomaly detection. %~\cite{pimentel2014review}. 
The most common approach to detect anomalies consists in modeling the distribution of normal samples by means of a background model of randomness. The design of this model is usually problem-specific. For instance, %in~\cite{du2010random} the model is built on the Fourier domain, 
in~\cite{du2010random} the normal samples are assumed to follow a Gaussian distribution, or in~\cite{text} the authors characterize normality by fitting different GMMs with dense covariance matrix to a 4-level image pyramid. As previously mentioned, the {\em a contrario} approach is based on the same principle, but differs from the others in the sense that it allows to derive detection thresholds by estimating and controlling the expected number of occurrences of an event under the background model. 
An excellent example is the work by Ehret et al.~\cite{ehret2019reduce}, in which they eliminate the self-similarity of the image by averaging the most frequent patches using Non Local Means~\cite{buades2005nlmeans}
, and perform a detection using the NFA score over the residual image, which is formed by noise and anomalies. In this way, the general background modeling problem gets reduced to a noise modeling problem, making the algorithm to work with any kind of background.

More recently, taking advantage of the continuously growing computation capabilities, new data-driven methods based on deep neural networks (DNN) were developed. In most cases they use only anomaly-free images, achieving impressive results in publicly available anomaly datasets. A comparison between some key elements of different state of the art works is presented in Table~\ref{tab:related-work}. From this table is easy to see that most of the methods operate with some kind of supervision, which is the main difference of our proposed algorithm. A common classification is to separate methods in two categories: \textit{feature-similarity based} and \textit{reconstruction based} methods. \\

\textbf{Feature-similarity based methods.} Works based on feature similarity start by building a new feature space, where it is claimed to be easier to identify anomalies. The second step is usually to define and measure some notion of similarity, which can range from a simple k-Nearest Neighbors (kNN) or some probability distribution fitting, to some learned metric, ad hoc to the specific problem. 
For example in~\cite{spade}, the authors construct a pyramid of features using the activation maps of a Wide-ResNet50, pre-trained on ImageNet, and use these feature maps for finding i) the K nearest anomaly-free images and ii) the pixel level anomaly segmentation. For the last one, they perform a multi-image correspondence of sub-images, matching the target sub-image with all sub-images of the K nearest neighbors. The anomaly score is finally obtained by averaging the distance between the sub-image of the target image and each one of the K neighbors. In~\cite{padim} excellent results are reported by modeling normality with a multivariate Gaussian distribution for each patch position, using the output of a pre-trained CNN, and measuring the Mahalanobis distance to normality for each patch. Normality is characterized for each patch using the output of a pre-trained CNN. At test time, to each patch of a test sample an anomaly score is assigned based on the Mahalanobis distance to normality for the same corresponding spatial region (patch). 
In \cite{fcdd} a fully convolutional neural network is trained, where the output features preserve the spatial information, and are indeed a down-sampled version of an anomaly heatmap. By using an HSC (Hyper-Sphere Classifier) loss, the nominal samples are encouraged to be mapped near the center of the new space, and the anomalous samples away. The network is trained with both normal and anomalous samples. Anomalous samples could be synthetic, but authors found that using even a few examples of real labeled anomalies the method perform much better. In~\cite{dict} the authors extract the features from a pre-trained DNN and build a dictionary over the features, which is subsequently used to measure the distance of the target image to normality. A student-teacher framework is used in~\cite{student-teacher}, where the student learns the distribution of anomaly-free images by matching their features with the teacher network accordingly. At test time, the feature maps of the two networks for a given target image is compared, and anomaly score obtained from their distance. In \cite{patch-svdd} the idea of \textit{Deep-SVDD}~\cite{deep-svdd} is extended by constructing a hierarchical encoding of the image patches. The \textit{Deep-SVDD} idea of mapping normal images near the center of an hyper-sphere in the feature space is improved by training an encoder to gather semantically similar patches. The output anomaly map is generated by computing the distance of each patch of the target image, to the nearest normal pre-computed patch in the feature space. The score for each pixel is the average of the scores of all patches it belongs to. The algorithm performs multi-scale inspection, and the results are aggregated by element-wise multiplication to generate the final anomaly map. 
The authors show that the performance of the whole algorithm is improved by adding a self-supervised learning term to the loss function. % As explained in~\cite{igd}, t
The hard boundary of SVDD associated with the hyper-sphere can cause the model to overfit the training data. Therefore, in~\cite{gsvdd} the authors propose to replace it by a Gaussian SVDD (GSVDD), and treat its mean and co-variance as latent variables to be estimated. They try to reduce the likelihood of an anomalous image embedding laying between normal samples by interpolating the latent embedding of different input images, making the distribution of the normal samples denser. The GSVDD is driven by the adversarially constrained auto-encoder interpolation (ACAI)~\cite{acai}, and by doing so, generates a robust estimation of the normal image distribution.

\textbf{Reconstruction-based.} Generative models, as VAEs~\cite{kingma2013auto} or GANs~\cite{goodfellow2014generative} are based on the idea that the network trained only on non-defective samples, will not be able to reconstruct the anomalies at test time. In practice, they usually present high capacity and generalization power, leading to even a good reconstruction of the anomalies. In~\cite{intra} the authors address this issue by stating the generative part as an inpainting problem. By building a network only with self-attention blocks, they are able to model successfully distant contextual information, and achieve good results. 
As stated in~\cite{bergmann-ssim}, better results could be obtained measuring the reconstruction error with the structural similarity measure (SSIM). Authors show that they achieve better results compared to classical $l^2$ distance. 
In~\cite{cutpaste} the authors design a special classification proxy task in order to be able to learn deep representation in a self-supervised manner, by training the network with random cut and pasted regions over anomaly-free images. Once the representation is obtained, a one-class classifier is trained using the new representation. 
In~\cite{dfr} the authors propose to train an Auto-Encoder over the feature maps of a pre-trained network. Instead of reconstructing the image itself, the network performs a deep features reconstruction. Finally they compute a pixel-level anomaly score by analyzing the difference between the features at the input and output of the Auto-Encoder. 
Unlike AEs and VAEs, which learn a direct representation from the image to the latent code, GANs learn the mapping from the latent space to realistic normal images. As the latent space transitions are smooth, in~\cite{anogan} the authors propose an iterative method to find a latent code that generates a similar image to the target, enabling to find anomalies by looking at the reconstruction error, using a Generative Adversarial Network.

When learning from one-class data, a usual approach consists in mapping normal images near some centroid in the latent space, and expect (but often with no guarantee), that an anomalous image will lay far from this centroid. There are many works that propose strategies overcome this potential issue. Many of them create some proxy task in order to train a self-supervised network. Even through the specific proxy task could be very different from the final objective, in some cases the network is proven to be very successful in extracting useful and informational features from the data. For example, in~\cite{rotnet}, the authors present a distribution-augmented contrastive learning technique via data augmentation, that obstructs the uniformity of contrastive representations, making it easier to isolate outliers from inliers. A deep neural network is trained over the anomaly-free images in a self-supervised scheme, by predicting the degree of rotation augmentation as the proxy task. In a second stage, a one-class classifier is built over the learned  high level data representation in order to obtain an anomaly score. \\

\begin{figure*}[t]
    \subfloat{\includegraphics[width=1\linewidth]{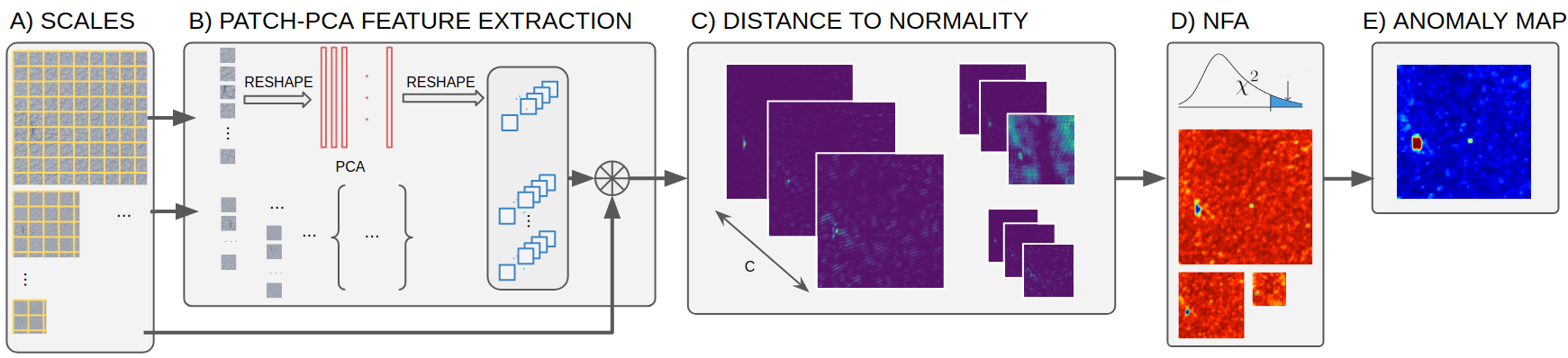}} \hfill
\\%[-20px]
    \caption{Method diagram. Our method can be divided in 5 stages. Stage A) consists in creating a pyramid of images of different scales. In stage B) we obtain the filters by performing a Patch-PCA transformation, and use them to filter the input image. At the beginning of stage C) we have one feature map for each filter and for each input channel, and we combine them by computing one mahalanobis distance for each channel. Stage D) performs the computation of the Number of False Alarms. The Stages B), C) and D) are computed for each scale independently, and all results are merged in stage E).}
    % \vspace{-10pt}
    \label{fig:method-diagram}
\end{figure*}

As we are specially interested in methods that can be used with no training and over a wide variety of different images as input, we also focus on the domain of \textit{saliency detection}. For this work we are only interested in textured images, where the images are uniform. In these cases, from the perception point of view, any structure or pattern that deviates from normality, should also be the most salient part of the image. There is an extensive literature dedicated to it. To cite a few relevant works, a united approach to the activation and normalization steps of saliency computation is introduced in~\cite{gbvs-harel2007graph}, by using dissimilarity and saliency to define edge weights on graphs that are interpreted as Markov chains. In~\cite{drfi-jiang2013salient} the salient object detection is formulated as a regression problem in which the task is learning a regressor that directly maps the regional feature vector to a saliency score. %This approach mainly considers regional contrast and backgroundness features, which makes it susceptible to failure in cluttered scenes. 
In~\cite{salicon-huang2015salicon}, the authors propose to estimate saliency by combining multiple popular DNN architectures for object recognition %(AlexNet, VGG-16, and GoogLeNet) 
that are known to encode powerful semantic features. They found that a good compromise between semantic representation and spatial information for saliency prediction is to use the last convolution layer. \\

Although there is a vast literature related to anomaly detection, we found that only few algorithms are able to perform the detection with no prior knowledge of the specific task/problem, and most of them try to model normality from a set of normal images, often needing them to be aligned and acquired in very controlled situations. In this work we develop a simple and effective unsupervised algorithm that only uses the input image, with no need of any prior knowledge or training data, and estimates a ``rareness score'' given by the Number of False Alarms (NFA) for each pixel/region of the image. This algorithm was developed to address a specific task, but it has proven to work well in a wide variety of different problems, with no modifications.

% ------------------------------------------------------------------
\section{Method}
\label{sec:method}
% ------------------------------------------------------------------

The kind of anomalies we wish to detect could be very diverse. In this work, we concentrate in low-level anomalies (by opposite of semantic anomalies). Using classic image processing techniques we achieved a simple, fast, and yet very effective method that has proven to work successfully in very different kinds of images, as will be shown in Section~\ref{sec:results}. As this method was developed to solve an industrial problem, both accuracy and speed are of major importance.\\

A key and reasonable assumption of our approach is that low-order moments of the set of normal samples can be statistical characterized from the whole dataset, i.e. that anomalies are so rare that do not affect these statistics. Under this assumption, the anomaly detection method we propose relies on a statistical characterization of the normal samples from the whole dataset, which we call {\em normality}, and a statistical distance of test samples to normality. This distance will then be used to evaluate how likely an observed sample is normal, and to derive a statistical test to detect anomalies.

Samples are characterized by a set of features that will be extracted to be uncorrelated. We show that these features follow a Gaussian distribution, which (assuming independence) allows us to easily define distance to normality as a Mahalanobis distance, and to ensure that this distance follows a $\chi^2$ distribution. 

The proposed method therefore consists of five stages: A) Image pyramid generation by scale decomposition, to detect anomalies of different sizes; B) Feature Extraction; C) Mahalanobis distance to normality; D) NFA computation; and E) Generation of the final anomaly map, by merging the results obtained for each different scales. 

For the feature extraction and the NFA computation (stages B and C), different variants are explored. In the remainder of this section we describe all the stages of the proposed approach.

% ------------------------------------------------------------------
\subsection{Image decomposition: scale's pyramid}
Since anomalies can occur at any scale, it is crucial to detect them no matter their size. This is performed by means of a multi-scale approach based on an image pyramid where each new scale is half the width and the height of the previous one. 

% ------------------------------------------------------------------
\subsection{Feature Extraction}

Three alternatives for the feature extraction stage are considered. \\

%entonces la salida de la etapa B tiene q ser features lo mas independientes posible
%Para ello consideramos 2 métodos, patch pca y resnet+pca. además, en la sección de resultados vamos a incluir un tercer método de extracción de caraterísticas: Gabor filters. Si bien estos filtros no cumplen con la condición de independencia, presentan algunas otras ventajas, como el tiempo de compjuto y q anda bien\\

\noindent
{\bf Patch-PCA.} The procedure described here is applied to each scale obtained from the scale's pyramid decomposition. This method is summarized in Figure~\ref{fig:method-diagram}.

To represent the local statistics of the pixels in the acquired images, such as its surrounding texture, we associate to each pixel a patch of $s \times s$ pixels centered on it. We treat each image channel separately, so for each channel $c$, each pixel location is characterized by a $n$-feature vector, where $n = s \times s$. %In this way, since each image consists of $c$ channels, each pixel location is characterized by a $n$-feature vector, where $n = c \times s \times s$. 
We apply PCA to decorrelate the patch feature vector and treat the new coordinates independently.   
%all coordinates and treat them independently.

PCA performs a linear transformation, $ \mathbf{y}_i = P^T \mathbf{x}_i$, where $\mathbf{x}_i \in \mathds{R}^n$ and $i \in \{1\dots N\}$ is a set of observations (patches), $ \mathbf{y}_i \in \mathds{R}^m$ with $m\leq n$ their projections, and the columns $\{\mathbf{p}_i\}_{i=1}^n$ of $P$ the PCA eigenvectors. The element $j$ in vector $\mathbf{y}_i$ is the result of the inner product between $\mathbf{x}_i$ and $\mathbf{p}_j$. Since these projections are inner products over patches, they can be efficiently computed as 2D convolutions with the eigenvectors as kernels, using common deep learning libraries. Extracting the PCA eigenvectors can also be seen as a methodology for finding the filters used to extract features, based only on the input image. This characteristic is of crucial importance in our application, as it defines specific adapted filters for each image. Once we have obtained the filters, we convolve the input image by channel with each filter (e.g. project into the PCA space), obtaining a feature vector of size $m$ for each of the $c$ channels, for each pixel. \\

\noindent
{\bf ResNet feature maps}. Instead of applying PCA to the output of the scale's pyramid, the PCA could be applied to the activation maps resulting from feeding a convolutional neural network with the input image. The advantage of this approach is that these activation maps gather a set of salient features of the image.  More precisely, the set of features we consider are obtained from a ResNet-50 network pre-trained on ImageNet, keeping the output of the last convolution of three layers:  \textit{layer1/conv3, layer2/conv3, layer3/conv3}. As the input of the network is the entire three-channel image, the filtering is not performed separately for each channel (as in the Patch-PCA approach), and at the end of this stage we obtain a feature vector of size equal to the sum of the number of filters in each convolutional layer that was considered. After applying PCA to the filtering output, we also keep the first $m$ components. \\

\noindent
{\bf Gabor filters.} Instead of performing the scale decomposition, and computing the Patch-PCA descriptors per image, we propose to use a set of (fixed, signal independent) Gabor filters of different sizes. Note that in this case the resulting features are not uncorrelated. However, results presented in Section~\ref{sec:results} show that this modification still achieves a good performance, while significantly reducing the computational burden from the image-dependent PCA computation. In this case we follow the same procedure than in Patch-PCA: the extracted features are the filters responses for each channel of the input image. %\textcolor{magenta}{In this case, the feature vectors that will be then used to compute the distance to normality are directly the output of the Gabor filters.}\\

\subsection{Distance to normality}
\label{sec:mahalanobis}

As previously mentioned, at this point we are supposed to have a set of uncorrelated features. This is true for the case of Patch-PCA and ResNet feature extractor, in the latter thanks to the PCA performed over the activation maps. For the case of Gabor filtering, since these filters are not orthogonal to each other, the resulting features do present some degree of correlation; nevertheless, we take the dare to analyze it anyways, as we believe it is still a valuable approach due to its simplicity. 
%this variant has the advantage of the simplicity and still achieves good results (that will be shown in Section~\ref{sec:results}).} 
%Also, avoiding the PCA computation makes this variant of the method faster, which could be of major importance, depending on the application.}

For each pair of features $(\mathbf{a},\,\mathbf{b})$, we define the distance between its corresponding image regions $\mathbf{x^a}$ and $\mathbf{x^b}$ as:
\begin{equation*}
    d^2(\mathbf{x^a}, \mathbf{x^b})  = \sum_{i=1}^m (a_i-b_i)^2.
\end{equation*}
To define a distance to normality based on the previous metric, we need to characterize the set of normal samples and their corresponding feature vectors (obtained with any of the feature extraction methods described in previous section). Assuming that the size of the anomaly is small compared to the size of the image, the low-order statistics of the normal samples can be accurately characterized from the whole image. In what follows we will describe the computation of the distance to normality for the features extracted with Patch-PCA and Resnet methods, since both end with uncorrelated feature vectors due to the use of PCA. Then, at the end of this section we will comment the case of Gabor feature extraction.\\

\begin{figure}[t]
\captionsetup{position=bottom}
    \subfloat
        [Patch-PCA features, keeping the first $m=3$ components.]
        {\includegraphics[width=\columnwidth]{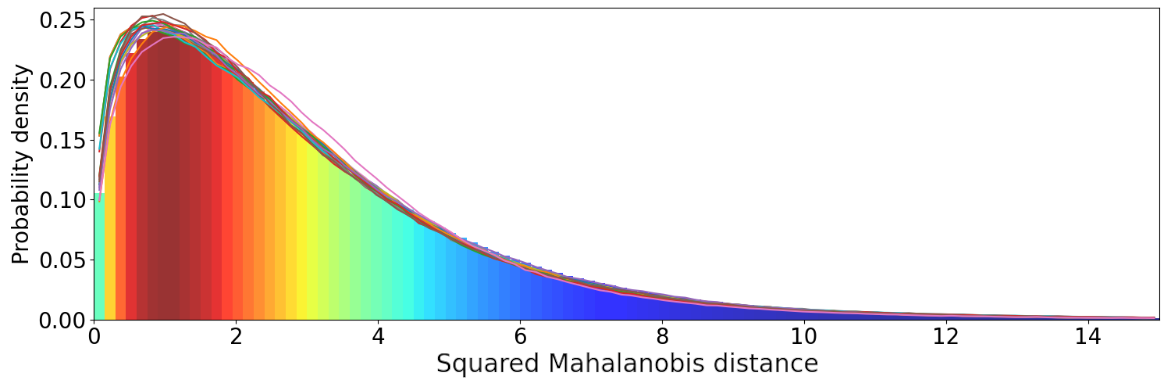}} \\
    \subfloat
        [ResNet-based features, for the first $m=5$ components of the PCA.]
        {\includegraphics[width=\columnwidth]{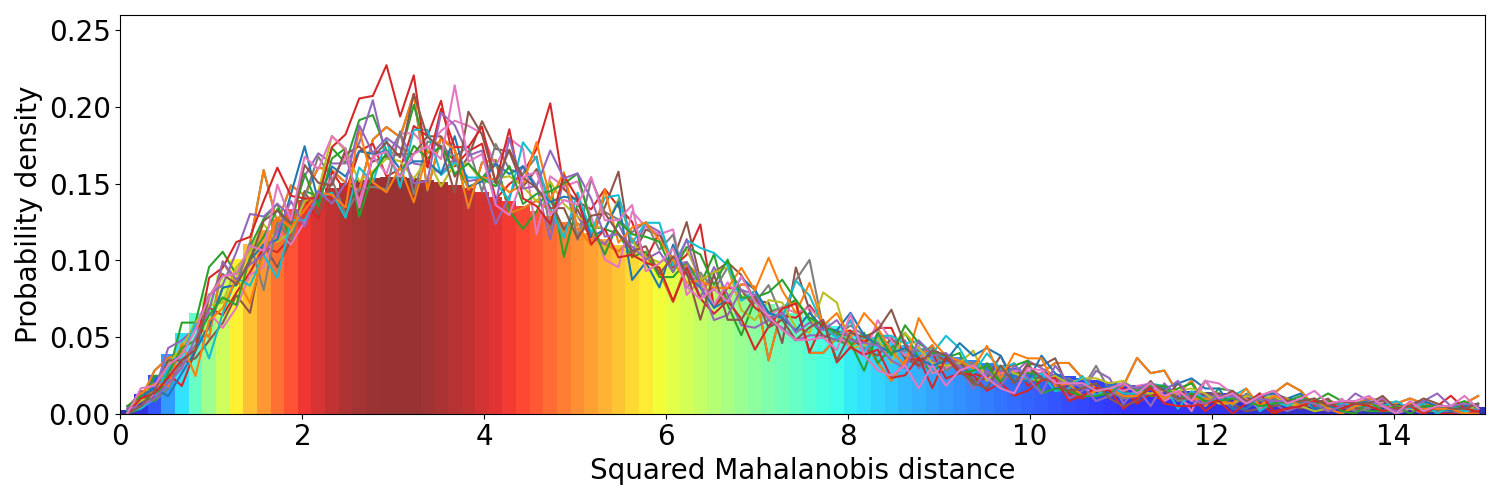}} \\
    \caption{Distribution of distances to normality for the Patch-PCA and ResNet-based features. Colored bars represent the $\chi^2$ theoretical distribution. Solid colored lines are the normalized histograms of distances to normality for different images.}
    \label{fig:chi}
\end{figure}

Let $\mu_i$ and $\sigma^2_i$, $i=1,\dots,m$, denote the mean and the variance of the top $m$ PCA components of the feature vectors. We define the squared distance of a sample to normality as:
\begin{equation}
   d^2(\mathbf{x^a}, \textit{normality}) = \sum_{i=1}^m \frac{(a_i - \mu_i)^2}{\sigma_{i}^2}.
   \label{ec:mahalanobis_dist}
\end{equation}
This quantity corresponds to the Mahalanobis distance of the sample $\mathbf{x^a}$ to the mean of the normal samples, restricted to the top $m$ principal directions. 
One of the main advantages of this simple notion of distance to normality is that, if we now consider samples $\mathbf{x^a}$ to be random, its statistical characterization is straightforward. Indeed, we have tested empirically that each PCA component follows a Gaussian distribution; knowing that these components are uncorrelated, if we further assume they are independent, the distance to normality \eqref{ec:mahalanobis_dist} follows a $\chi^2$ distribution with $m$ degrees of freedom.
Figure~\ref{fig:chi} shows the distribution of distances from image samples to normality for several test images, and its comparison to the $\chi^2(m)$ distribution. Note that the fit is remarkably accurate.

In the case of the method based on Gabor filters, for simplicity, we will assume uncorrelated components and apply the same computation of the distance to normality explained above. Although not true, as already explained, does not affect the results while allowing a computational efficient implementation.

\subsection{Number of False Alarms}
\label{sec:nfa}

In this section we propose three different strategies, sharing a common methodology, to detect anomalies in images of textures. The detection methodology we follow is the {\em a contrario} framework. It is a multiple hypothesis testing strategy that detects meaningful events (image structures) as large deviation from a background model or null hypothesis ${\mathcal H}_0$, which in our case models normality. Following the a contrario approach~\cite{desolneux2007gestalt}, to each event $\mathcal E$ we inspect, we associate a rareness score called Number of False Alarms (NFA) defined by
$$\text{NFA}({\mathcal E}) = N_{\text{tests}} \mathbb{P}_{{\mathcal H}_0}({\mathcal E}).$$
$N_{\text{tests}}$ is the number of events we are theoretically testing, and $\mathbb{P}_{{\mathcal H}_0}$ is the probability of observing ${\mathcal E}$ under ${\mathcal H}_0$. An event $\mathcal E$ is said to be $\varepsilon$-meaningful if $\text{NFA}({\mathcal E}) < \varepsilon$. It is easy to show that, under the background model, the expected number of $\varepsilon$-meaningful events is lower than $\varepsilon$~\cite{desolneux2007gestalt}. In other words, the lower the NFA of an event is, the more unlikely that this event was produced by the background model, being more likely to result from another cause. If we set the threshold on the NFA to $\varepsilon = 1$, we should detect at most one false alarm, and the rest of the detections should correspond to actual anomalies. 

We now present three different detection strategies based on the a contrario framework. It is usually the case where the anomaly only shows in one particular channel or scale (e.g. some color anomaly). Therefore, in order to keep all possible detections, we keep the minimum of all NFAs across all channels and scales. The output of this stage is one NFA map for each scale. \\

\noindent
{\bf NFA by pixels} In this strategy, for each pixel $\mathbf{x^a}$ in the image, characterized by its feature vector $\mathbf a$, we define 
$$
\text{NFA}({\mathbf{x^a}}) = HW (1-\mbox{CDF}_{\chi^2(m)}(d^2(\mathbf{x^a},\,normality))).
$$
This quantity corresponds to the number of events we test (all the pixels in the image) times the probability of a $chi^2(m)$ random variable being larger than its observed squared distance to normality. \\

\begin{table*}[t]
% \vspace{-15pt}
\caption{\vspace{2pt}Area under the receiver operating characteristic curve (ROC AUC). In red the best score, blue the second best and green the third. The ranking of the algorithms is done by sorting the performance and summing the positions for each algorithm. The one with lowest value is the best. For our methods, the $\dagger$ indicates the Block NFA variant, the $\ddagger$ indicates the Region NFA variant, and no symbol corresponds to the Pixel NFA variant.}\label{tabresultsmvtec}
% \vspace{-5pt}
\centering
    \begin{tabularx}{\linewidth}{X X X X X X X X X X X X X X X X}
\toprule
\textbf{Method} & AE SSIM \cite{bergmann-ssim} & AE L2 \cite{bergmann-ssim} & Ano Gan \cite{anogan} & Cnn Dict \cite{dict} & Tex Insp \cite{text} & DRFI \cite{drfi-jiang2013salient} & GBVS \cite{gbvs-harel2007graph} & Salicon \cite{salicon-huang2015salicon} & Det Noise \cite{ehret2019reduce} & Ours PCA & Ours PCA$^\dagger$ & Ours Gabor & Ours Gabor$^\dagger$ & Ours ResNet & Ours ResNet$^\ddagger$\\ 
\midrule

\textbf{Carpet} &  \textcolor{green}{\textbf{0.87}} &  0.59 &  0.54 &  0.72 &  \textcolor{blue}{\textbf{0.88}} &  0.72 &  0.45 &  0.53 &  0.57 &  0.76 &  0.65 &  0.77 &  0.69 &  \textcolor{red}{\textbf{0.94}} & \textcolor{red}{\textbf{0.94}} \\

\textbf{Grid} &  \textcolor{blue}{\textbf{0.94}} &  0.90 &  0.58 &  0.59 &  0.72 &  0.58 &  0.73 &  0.79 &  0.68 &  0.88 &  0.91 &  0.90 &  0.83 &  \textcolor{green}{\textbf{0.92}} & \textcolor{red}{\textbf{0.96}} \\

\textbf{Leather} &  0.78 &  0.75 &  0.64 &  0.87 &  \textcolor{green}{\textbf{0.97}} &  0.68 &  0.86 &  0.76 &  0.81 &  \textcolor{blue}{\textbf{0.98}} &  0.87 &  0.95 &  0.92 &  \textcolor{red}{\textbf{0.99}} & \textcolor{red}{\textbf{0.99}} \\

\textbf{Tile} &  0.59 &  0.51 &  0.50 &  \textcolor{red}{\textbf{0.93}} &  0.41 &  0.72 &  0.49 &  0.40 &  0.53 &  0.69 &  \textcolor{green}{\textbf{0.84}} &  0.79 &  0.80 &  0.77 & \textcolor{blue}{\textbf{0.87}} \\

\textbf{Wood} &  0.73 &  0.73 &  0.62 &  \textcolor{blue}{\textbf{0.91}} &  0.78 &  0.67 &  0.79 &  0.72 &  0.57 &  0.86 &  \textcolor{green}{\textbf{0.87}} &  \textcolor{green}{\textbf{0.87}} &  0.84 &  0.86 & \textcolor{red}{\textbf{0.92}} \\

\midrule

\textbf{Mean} &  0.78 &  0.70 &  0.58 &  0.80 &  0.75 \  &  0.67 &  0.66 &  0.64 &  0.63 \  &  0.83 &  0.83 &  \textcolor{green}{\textbf{0.86}} &  0.82 &  \textcolor{blue}{\textbf{0.90}} & \textcolor{red}{\textbf{0.94}} \\

\midrule

\textbf{Ranking} & 8 & 10 & 15 & 6 & 9 & 11 & 12 & 14 & 13 & 5 & 4 & \textcolor{green}{\textbf{3}} & 7 & \textcolor{blue}{\textbf{2}} & \textcolor{red}{\textbf{1}}\\

\bottomrule
        \end{tabularx}
% \vspace{-10pt}
\end{table*}
% ------------------------------------------------------------------

\noindent
{\bf NFA by blocks} In this per-block strategy the objective is to detect anomalies as conspicuous accumulation of pixels exhibiting a large distance to normality within image blocks. This algorithm begins with the extraction of candidate regions for each channel, by thresholding each squared Mahalanobis distance $d^2(\mathbf{ x^a},\,normality)$ with a threshold $\tau$, which can be easily set. Since under the normality hypothesis, these distances are assumed to follow a $\chi^2(m)$ distribution, we set $\tau$ such as the corresponding $p$-value is $p=0.01$. In this case, our normality assumption ($\mathcal{H}_0$), is that pixels such that their distance to normality exceeds $\tau$ are uniformly distributed on the whole image. Following the idea in~\cite{marina}, we base this NFA criterion on the detection of high concentrations of these candidates as follows.

Considering a block $B$ of size $w \times w$, and the set of candidates $\mathcal{L}_B=\{\mathbf{ x^a} \in B \text{ s.t. } d^2(\mathbf {x^a},\,normality) > \tau\}$ within the block, we define a new random variable, $U_{i,j}$, with the following Bernoulli distribution: %$U_{i,j}$ that takes value $1$ if $(i,j) \in \mathcal{L}_B$ and $0$ otherwise. 
$$
U_{i,j} = 
     \begin{cases}
       1 &\;\text{if }(i,j) \in \mathcal{L}_B\\
       0 &\;\text{otherwise.}\\
     \end{cases}
$$

To search for unusual concentrations of candidate pixels, we need to evaluate the probability %$P(U \geq |\mathcal{L}_B |)$, with $U=\sum_{i,j} U_{i,j}$. 

$$P(U \geq |\mathcal{L}_B |),\;\text{ with } U=\sum_{i,j} U_{i,j}\text{ following a binomial law}.$$

Evaluating this probability is not straightforward, as the $U_{i,j}$ are not independent. To overcome this issue we would have to sub-sample the Mahalanobis distance image, using a factor that ensures that two neighbor pixels in the new image can be considered independent, i.e. keeping one of each $s$ pixels in each direction (the size of the patch for the Patch-PCA feature extraction, or the size of the filter for Gabor filtering). Once we have the sub-sampled image and we have independent pixels, we could use the previous NFA by pixel approach. But instead of doing that, we can use a trick based on the intuition that we are observing $|\mathcal{L}_B|^2/s^2$ independent meaningful events, from a total of $w^2/s^2$ possibilities. Note that if were actually doing the sub-sampling, we probably would not find the exact average value ($|\mathcal{L}_B|^2/s^2$), depending on the grid origin we choose. In fact, we will only find this value if the meaningful events are uniformly distributed. In the case they are not, we would obtain some sub-samplings where we find more events than the average, and some others where we find less. In any case, we can ensure that by choosing the average value we will have at least one sub-sampling as significant as the one we are evaluating. Therefore, %, we consider $s^2$ separate tests, one for each possible sub-sampled grid by a factor $s$ in both dimensions. For each one we observe $w^2/s^2$ distances. 
the probability of observing at least $|\mathcal{L}_B|/s^2$ distances greater than $\tau$ in block $B$, is given by the tail of the binomial law, $\mathcal{B} \left( \frac{w^2}{s^2}, \frac{|\mathcal{L}_B|}{s^2}, p \right)$. The number of blocks $B$ to be tested is $HW/w^2$. Testing all $s \times s$ sub-sampled grids per block leads to a total number of $N_T = HW s^2/w^2$ tests. Finally, the NFA associated to a block $B$ is given by
$$ \text{NFA} = \frac{HW}{w^2} s^2 \mathcal{B} \left( \frac{w^2}{s^2}, \frac{|\mathcal{L}_B|}{s^2}, p \right).$$

%$w^2/s^2$ the number of independent tests, $|\mathcal{L}|/s^2$ is the number of independent events,
\vspace{5pt}
\noindent
{\bf NFA by regions} As in the NFA by blocks strategy, here we also look for conspicuous accumulation of pixels for which the distance to normality is sufficiently large.

The previously presented NFA computation techniques apply to pixels and square blocks, but this variant extends the possibilities by enabling to compute the NFA of arbitrary sized and arbitrary shaped regions. It is inspired on the work by Grompone et al.~\cite{von2021ground}.
%, and also makes use of some of the ideas previously discussed in the ``NFA by blocks'' case.
One of the difficulties of computing the NFA value over regions of any shapes resides in the calculation of the number of tests. To do so, we need to quantify all possible arrangement of pixels with any shape, and of all possible sizes. To build these regions, Grompone et al. propose to work with 4-connectivity regions $\mathcal{R}$, for which the approximate number of possible configurations with $N_\mathcal{R}$ pixels is given by (see~\cite{jensen2000statistics}): $$\#\text{configurations}\approx \alpha \frac{\beta ^{N_{\mathcal{R}}} }{ N_{\mathcal{R}}},$$ where $\alpha = 0.316915$ and $\beta = 4.062570$. The construction of the regions is based on a greedy algorithm proposed in~\cite{von2021ground}, but with some differences. First, we use the Mahalanobis distance as input, and consider the pixels for which the distance to normality is larger than a fixed threshold. Secondly and more important, we add a region growing criterion based on the NFA value itself. In order to decide if a pixel is to be added to a certain region, we evaluate the significance (NFA value) of the region with and without the pixel, and only add it if it makes the whole region more significant, i.e. lowers the NFA value. The original algorithm is designed for single channel images, so we apply it for each filter response separately, and then combine all the NFAs by keeping the minimum between all results for each position. The pixels that were not considered in any region are assigned with the maximum NFA value calculated, as they are the ones less significant.

We implemented this variant of the NFA computation only for the ResNet-based feature maps, which is the best performing combination among the methods presented so far, as will be shown in Section~\ref{sec:results}.

As explained before, once we obtain the feature maps (in this case obtained only from \textit{layer1[0]/conv3}), we decorrelate them using PCA and compute the Mahalanobis distances. Note that the feature maps have a spatial resolution much lower than the input image. In order to work at full resolution, we up-sample the Mahalanobis distances to the input size and compute the NFA. By doing so, we are introducing some dependency between neighbor pixels. To overcome this issue, we proceed in way similar to the NFA by blocks strategy, by considering the receptive field $\tilde{s}$ of the network to ensure independence again. Reasoning in a similar way, by assuming that we observe $N_\mathcal{R} / \tilde{s}^2$ pixels, we can ensure that there is at least one event as significant as the one we are computing. The final NFA value computed is therefore:
\begin{equation*}
\begin{split}
    \text{NFA} & (\mathcal{R}) = \\
    & \frac{HW}{\tilde{s}^2}\, \alpha \frac{\beta ^{N_{\mathcal{R}}/\tilde{s}^2} }{ N_{\mathcal{R}}/\tilde{s}^2} \bigg( 1-\mbox{CDF}_{\chi^2(N_{\mathcal{R}} / \tilde{s}^2)} \Big( \frac{1}{\tilde{s}^2} \sum_{(i,j) \in \mathcal{R}} d^2_{i,j} \Big) \bigg),
\end{split}
\end{equation*}
where $d^2_{i,j}$ denotes the squared distance to normality associated to pixel $(i,j)$.
% ------------------------------------------------------------------
\subsection{Final anomaly map}
At this point, we have obtained one NFA map for each scale. The next step is to combine them in order to generate a unique map gathering the information of all scales. To do so, we follow the intuition that if some structure results to be anomalous at a certain scale, it has to be marked as anomalous in the final map. Therefore we compute the final NFA map as the minimum value of NFA of each scale for each pixel. These maps are combined by up-sampling smaller scales to match the original size. Finally, we translate this NFA map into an rareness measure by simply defining the Anomaly Score as $AS=-\log_{10} (\text{NFA})$, which provides a more intuitive map where the larger the value associated to a structure, the higher its  rareness.

\section{Results}
\label{sec:results}
% ----------------------------------------------------------------

%To the best of our knowledge, from the wide literature available for anomaly detection, there are few methods truly unsupervised, that can function looking only to one image, with no prior training or information about normal samples, as shown in Table~\ref{tab:related-work}. 

To the best of our knowledge, there are few truly unsupervised methods that can work only with the image being inspected, with no prior training or information about normal samples, as shown in Table~\ref{tab:related-work}. In order to compare our proposed method with the state of the art, we use MVTec AD~\cite{mvtec}, a recent anomaly dataset that simulates faults or anomalies in industrial conditions. The MVTec AD dataset contains several subsets, divided in two categories: texture and objects. We focus on the texture category, as we are interested in the detection of low level anomalies. Also, one of the subsets has special importance for us, 
%as it has images of the same kind as the industrial application that motivated this work:
as it considers anomalies in leather samples. MVTec AD also provides several non-faulty images, which most of the state of the art methods use for learning the normality. In order to extend the comparison, we also include the baseline methods, even if they require to train on normal-only samples and the comparison is not completely fair, in their favor. Results are shown in Table~\ref{tabresultsmvtec}. To perform the comparison, we use the area under the receiver operating characteristic curve (ROC AUC), as it is the most widely used in the literature, and enables us to compare with state of the art results.

\begin{figure}[t]
    \subfloat{\includegraphics[width=\columnwidth]{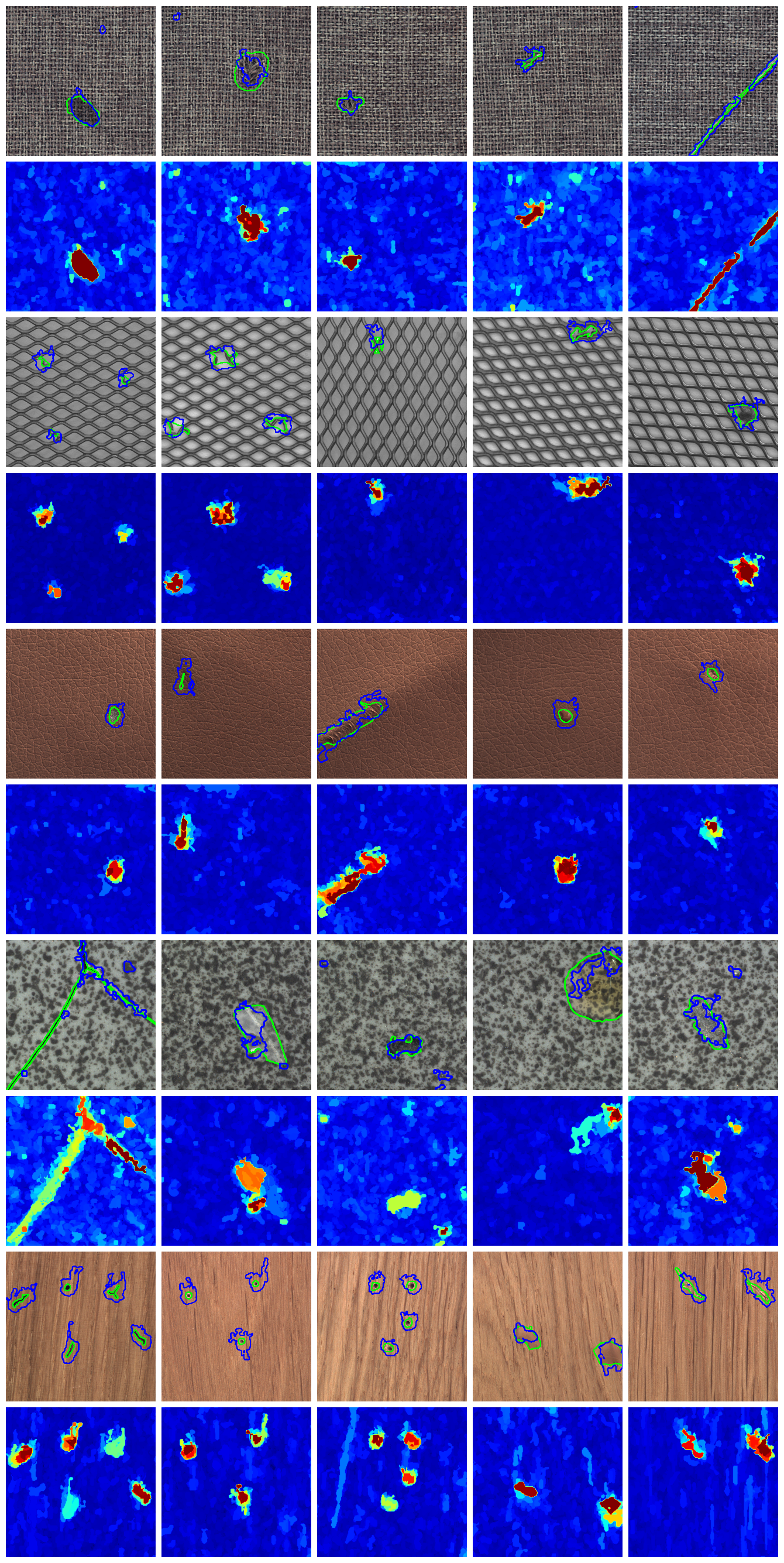}}
    \caption{For each textured dataset in MVTec AD (carpet, grid, leather, tile, and wood), we show one image with each type of defect and their corresponding anomaly maps with the \textit{ResNet+RegionNFA} method. The ground truth in is shown green, and the detection with log-NFA=0 in blue, superimposed to the original image.}
    \label{fig:results_mvtec}
\end{figure}

The results in Table~\ref{tabresultsmvtec} corresponding to the Block-NFA version are indicated with a cross; the others use Pixel-NFA. The proposed algorithm achieves state-of-the-art results, comparing it even with some methods that have a clear benefit by using anomaly-free images for training. In some subsets our methods manage to obtain the best results. As mentioned before, we pay special attention to the leather case since it is related to our industrial application (see next section). For this dataset, our approach achieves the highest score.
%, as it's 
%a closely related problem to our industrial
%our application, that will be presented in next section. 
Furthermore, if we consider the average score over all subsets, all variants of the proposed algorithm perform the best, among which stands out the ResNet variant, that in addition to produce the best results, it is also the fastest variant of our approach. In addition, the final improvement added to the ResNet version, which enables to compute the NFA for arbitrary shaped and arbitrary sized regions, has proven to outperform all the other methods, even being the best in almost all subsets. Figure~\ref{fig:results_mvtec} shows several example images alongside with their final anomaly map, for one example of each defect type of each dataset. Both the ground truth (in green) and the detections with log-NFA=0 (in blue) are drawn over the original image.

\begin{figure}
    \centering
    \subfloat{\includegraphics[width=\columnwidth]{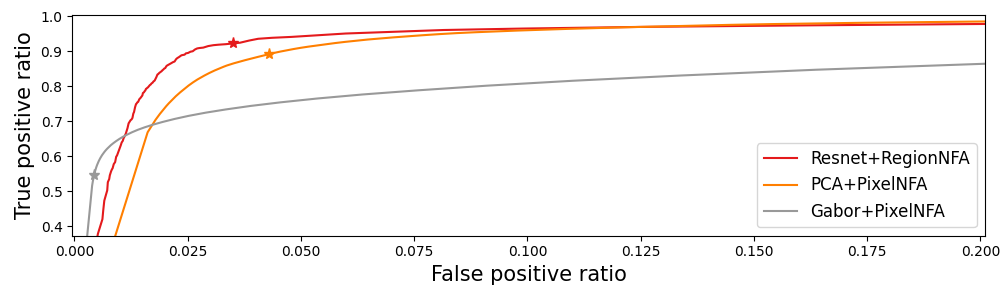}} \hfill
    \caption{ROC curve for the leather subset of MVTec AD, for the best performing variant for each feature extractor of the proposed method. The value with AS=0 (NFA=1) is indicated with a star. Note that the points AS=0 lay close to the optimal points of the ROC curves.}
    \label{fig:roc}
\end{figure}

\begin{figure}
    \hfill 
    \subfloat{\includegraphics[width=.96\columnwidth]{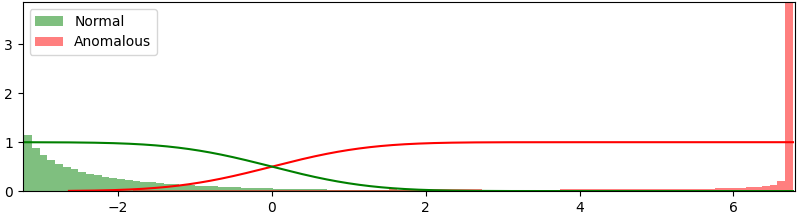}} \hfill     
    \caption{Distribution of the Anomaly Score $\text{AS}=-\log(\text{NFA})$ values for normal and anomalous pixels of the leather subset of MVTec AD. overlapped with the cumulative density function for both classes. Note the large gap between classes, and the adequate choice of NFA values close to AS=0 to derive the threshold.}
    \label{fig:gap}
\end{figure}

As mentioned before, one of the benefits of using the \textit{a contrario} framework is the simplicity to set the threshold for segmentation. As the metric we used to compare with all methods is independent of the threshold, we present the Figure~\ref{fig:roc}, where we show the curves obtained for the leather dataset with the top three performing variants of our proposed algorithm (for each different feature extractor), and indicate with a star the point where AS=0, i.e. NFA=1. It can be seen that values close to $\text{AS}=0$ lay near the optimal point in the ROC curve. Also, another interesting verification is to estimate the GAP between the anomaly scores of the faulty and non-faulty regions. In order to do so, we show in Figure~\ref{fig:gap} the distribution of the anomaly scores for all leather samples, overlapped with the cumulative density functions for both classes, making it clear that there is a good separation between them. Additionally, we evaluate this GAP numerically by computing the difference between the median values of the anomaly scores $\text{AS}$ of the two classes for all variants, obtaining the results shown in Table~\ref{tab:gap}.
\begin{table}
\normalsize
\begin{center}
\caption{\vspace{2pt}GAP between the anomaly scores of the faulty and non-faulty regions for the different combinations of our method.}
\label{tab:gap}
\begin{tabular}{l l}
    \toprule
    Method & GAP \\
    \midrule
    PCA+PixelNFA & 2.25\\
    PCA+BlockNFA & 4.06\\
    Gabor+PixelNFA & 9.12\\
    Gabor+BlockNFA & 4.62\\
    ResNet+PixelNFA & 3.15\\ 
    ResNet+RegionNFA & 11.83\\
    \bottomrule
\end{tabular}
\end{center}
\vspace{-12pt}
\end{table}
%\begin{itemize}
%    \item \textit{PCA+PixelNFA}: 2.25, %5.18, 
%    \item \textit{PCA+BlockNFA}: 4.06, %9.35, 
%    \item \textit{Gabor+PixelNFA}: 9.12, %21.00, 
%    \item \textit{Gabor+BlockNFA}: 4.62, %10.64, 
%    \item \textit{ResNet+PixelNFA}: 3.15, % 7.26. 
%    \item \textcolor{blue}{\textit{ResNet+RegionNFA}: 11.83.}
%\end{itemize} 
In all cases we obtained a very good GAP, and well separated classes. We can observe that depending on the feature extraction used, in one case the AS gap is greater using BlockNFA, and in the other using PixelNFA. The best GAP is obtained for the combination Resnet+RegionNFA, which is also the best performing method. 

For the PCA-based variant we used $m=45$, $s=17$, and 4 scales. For the Gabor variant $m=72$, $s$ ranging from 7 to 31, and 4 scales. The Block NFA is computed considering blocks of size $w=51$ with a stride of 10 pixels.

%\textcolor{red}{vale la pena agregar una frase sobre gabor? o la dejamos por ahí?}

% ------------------------------------------------------------------
\section{Anomaly detection in leather for the automotive upholstery industry}
\label{sec:bader}
% ------------------------------------------------------------------

In this section we present the industrial problem that motivates this work, carried out for a world leading producer of leather upholstery for the high-end automotive industry\footnote{The project was developed by Digital Sense in collaboration with Pensur.}.   %namely the detection of defects in leather pieces.

The raw skin of animals can present multiple kinds of defects, caused by the living animal itself (e.g. tick, lice and mite marks, scars), % warts, grubs, scars, thorn and horn scratches, neck wrinkles) %, veininess, structural differences, humps, brand marks)
or in the subsequent processes that transform raw leather into textured leather pieces ready to be installed in a vehicle. Various types of defects (holes, machinery drag, etc.) may be introduced in each step of the process, such as differences in the processes of tanning, differences on the roller pressing for texturing (Figure~\ref{fig:example-images}(a)-(b)), etc. The size and shape of defects vary widely (from bites,  which are millimeters in size, to wrinkles occupying most of the piece). In industries that work with leather, these defects must be detected as soon as possible in the production line to take the corresponding actions (avoiding cutting on the defect, leaving it in inconspicuous areas, or even discarding the pieces). Having a human operator to perform the quality control has several downsides, such as the inability to maintain the necessary attention level for long periods, criterion differences between different operators, and of course the time and workforce costs.

Each vehicle manufacturer allows a small margin of defects in the leather pieces that they buy. When this margin is exceeded the entire batch of leather pieces is returned, incurring in large losses for the producing companies. On the other hand, production is limited by the number of personnel available for inspection. Since defects are searched manually in all samples at various points in the process, this is a bottleneck in the production capacity of the plant.

% The skin samples considered in this work were provided by a leading international producer of premium leather for the automotive industry. 
As the defects are often very hard to see to the naked eye, a dedicated acquisition setup was designed. %in collaboration with XXX\footnote{Excluded due to double blind review}, an engineering company that provides industrial solutions. 
For each skin sample, a total number of 5 images are acquired with a single camera pointing at the sample at nadir, using different lightning conditions and orientations. More precisely, a diffuse light next to the camera and 4 directional grazing lights, located in each corner of the table. In this way, some defects that are very subtle but have little relief can also stand out. A typical set of images acquired with this setup are shown in Figure~\ref{fig:pca_lights_merge}, top row.

% ------------------------------------------------------------------
\subsection{Pre-processing} 

\begin{figure}
\centering
    \includegraphics[width=\columnwidth]{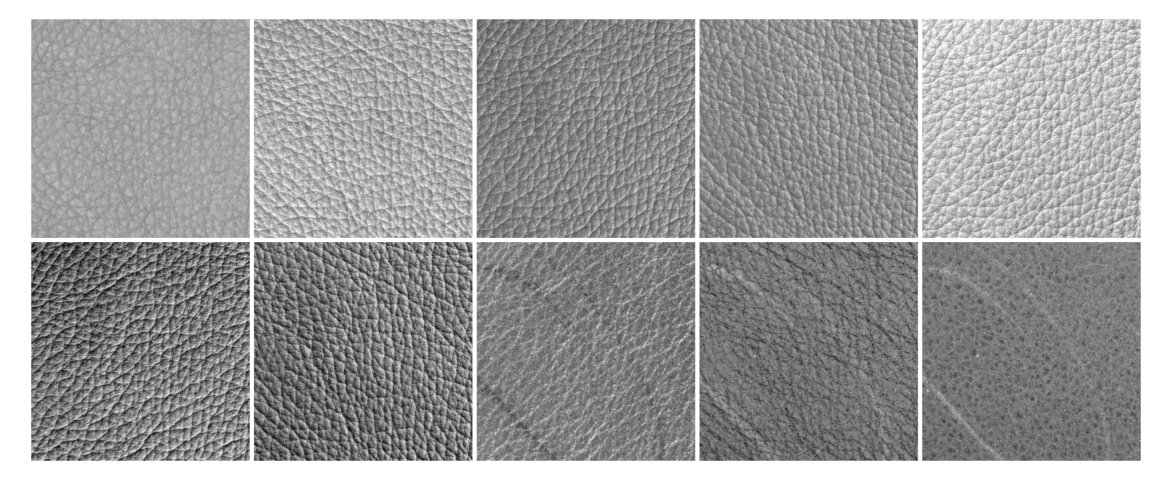}
    %\vspace{-10pt}
    \caption{Top row: original images. Bottom row: Output of the preprocessing step. Note how the PCA-based preprocessing better manages to single out potential anomalies.}
    %\vspace{-10pt}
    \label{fig:pca_lights_merge}
\end{figure}

The algorithms presented in Section~\ref{sec:method} can be fed either with all 5 raw images or with the images resulting from this pre-processing step. 
Contrarily to the most common use of PCA, where only the first components are kept, in this case we are more interested in the last ones. Indeed, the first components correspond to the directions of largest variance and therefore encode global information of the image, such as color or texture. Since anomalies are particular local structures, they are mostly present in the last components. Figure \ref{fig:pca_lights_merge} depicts a visual example. The first row corresponds to the 5 raw images, and the second row to the PCA components (the first image corresponds to the first eigenvector). This example shows that the anomaly is very hard to visualize in all the raw images, but stands out more clearly in the last 3 components of the projected space. %In this case, the component which makes the anomaly more visible is the third one.

\subsection{Industrial application results}
\begin{figure}[t]

\subfloat{\includegraphics[width=0.195\linewidth]{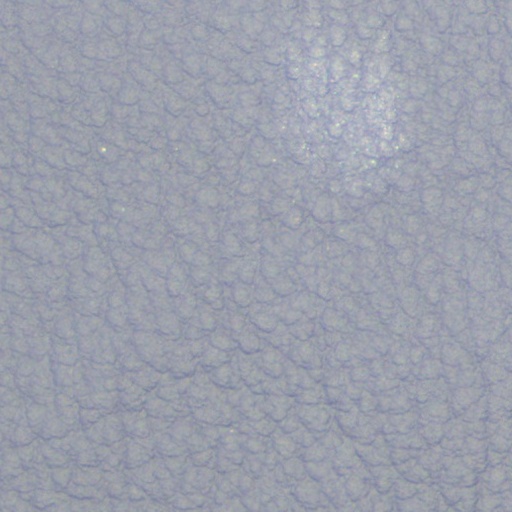}} \hfill
\subfloat{\includegraphics[width=0.195\linewidth]{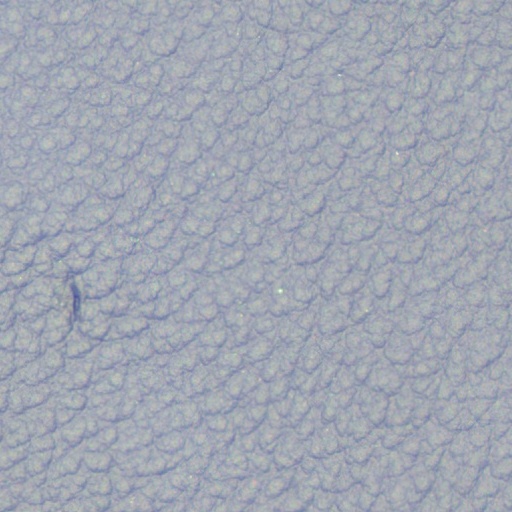}} \hfill
\subfloat{\includegraphics[width=0.195\linewidth]{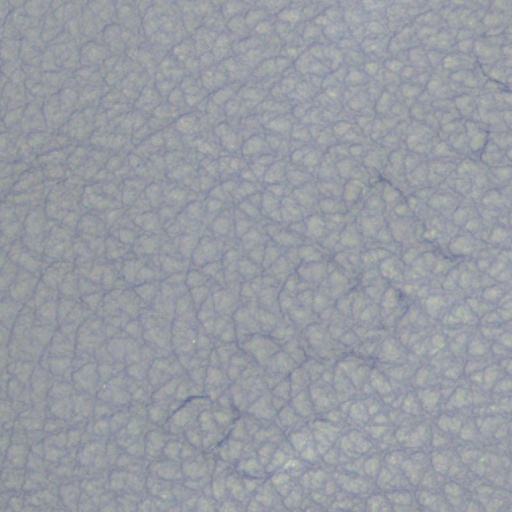}} \hfill
\subfloat{\includegraphics[width=0.195\linewidth]{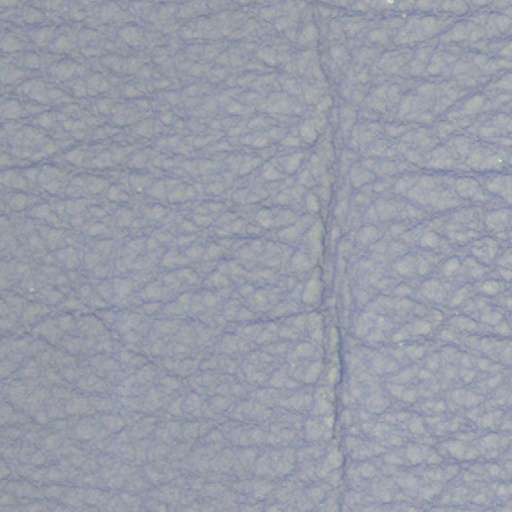}} \hfill
\subfloat{\includegraphics[width=0.195\linewidth]{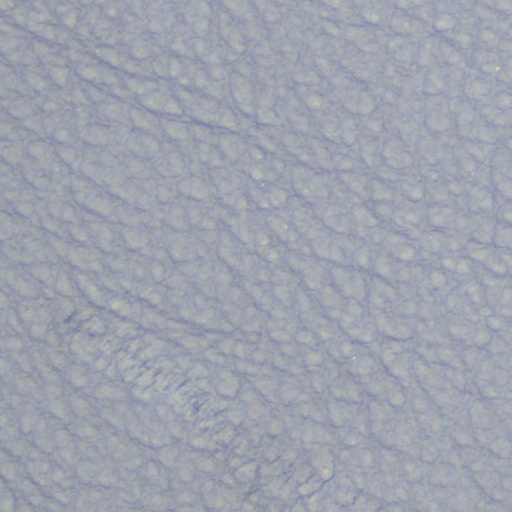}} \hfill \\[-20px]

\subfloat{\includegraphics[width=0.195\linewidth]{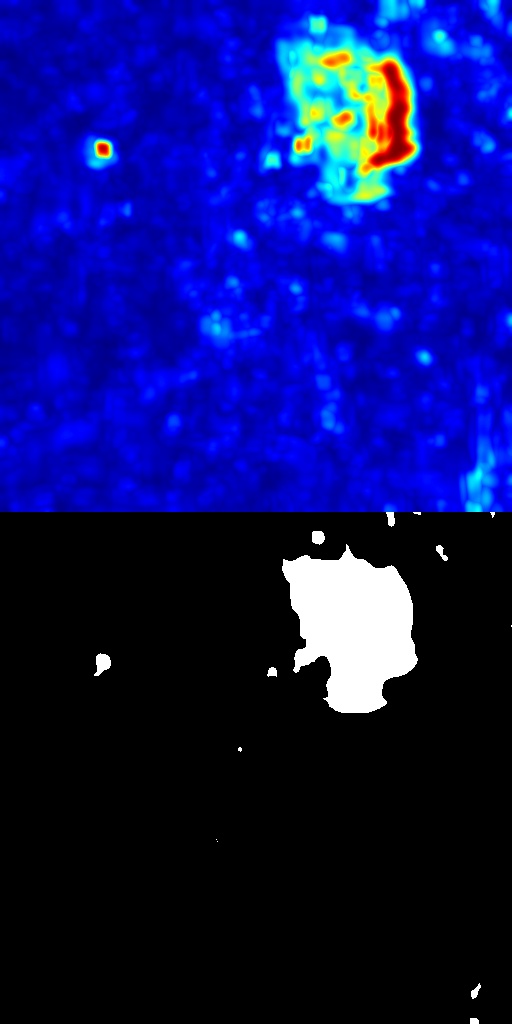}} \hfill
\subfloat{\includegraphics[width=0.195\linewidth]{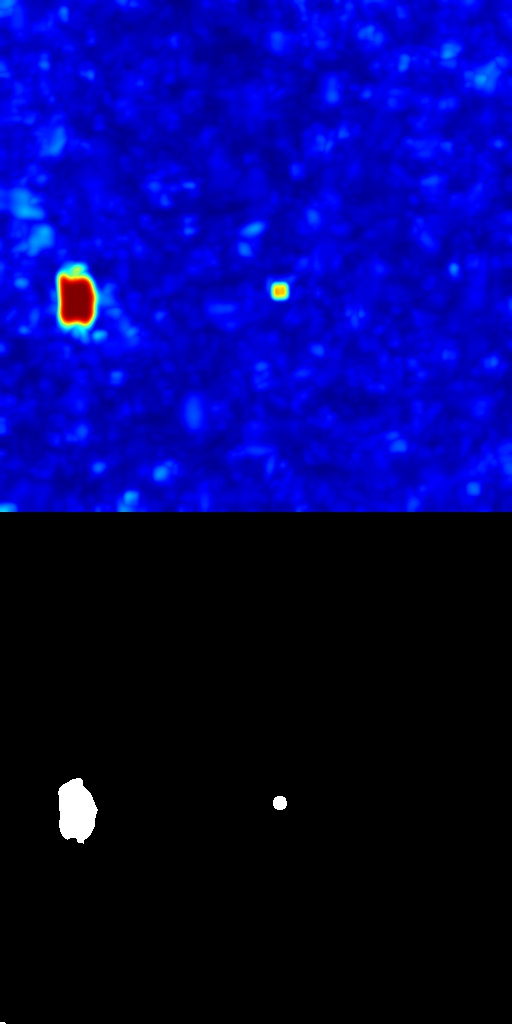}} \hfill
\subfloat{\includegraphics[width=0.195\linewidth]{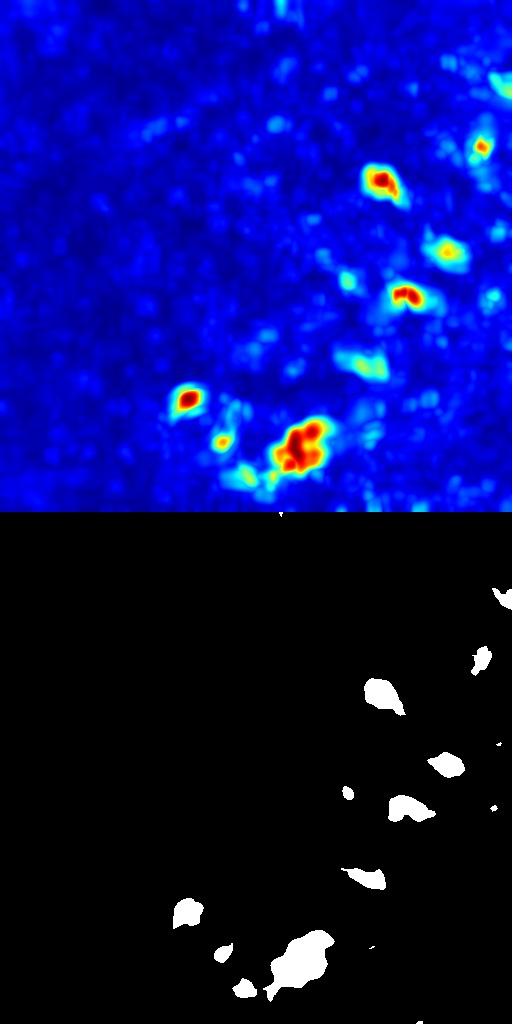}} \hfill
\subfloat{\includegraphics[width=0.195\linewidth]{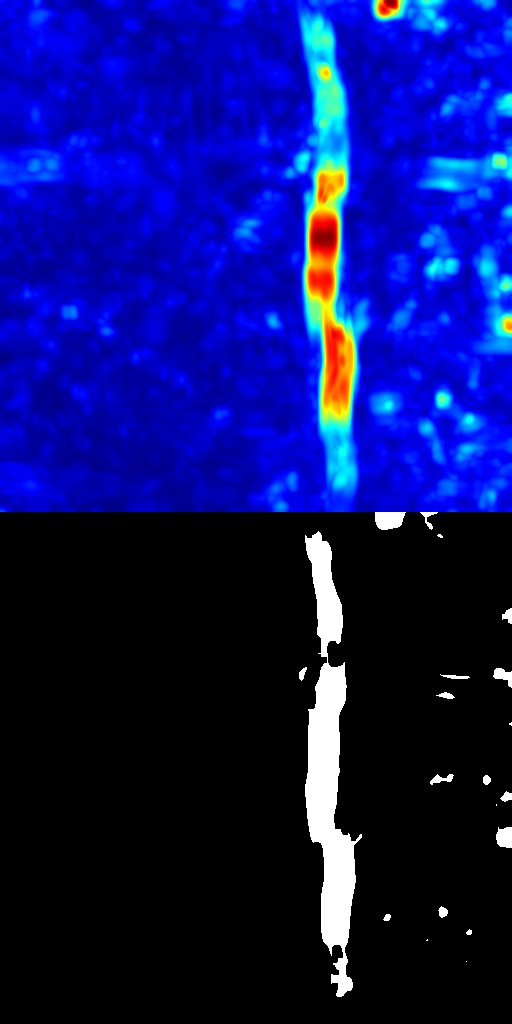}} \hfill
\subfloat{\includegraphics[width=0.195\linewidth]{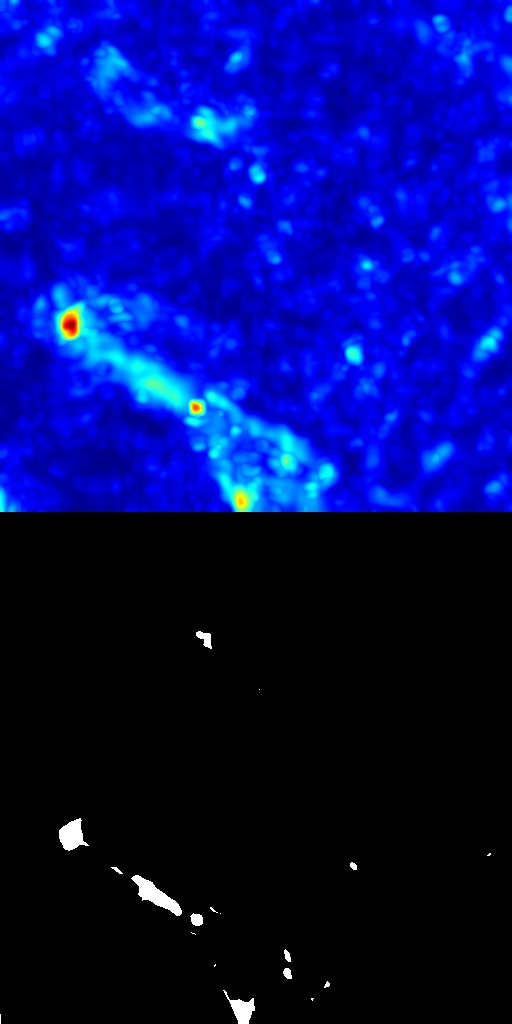}} \hfill \\[-20px]

\subfloat{\includegraphics[width=0.195\linewidth]{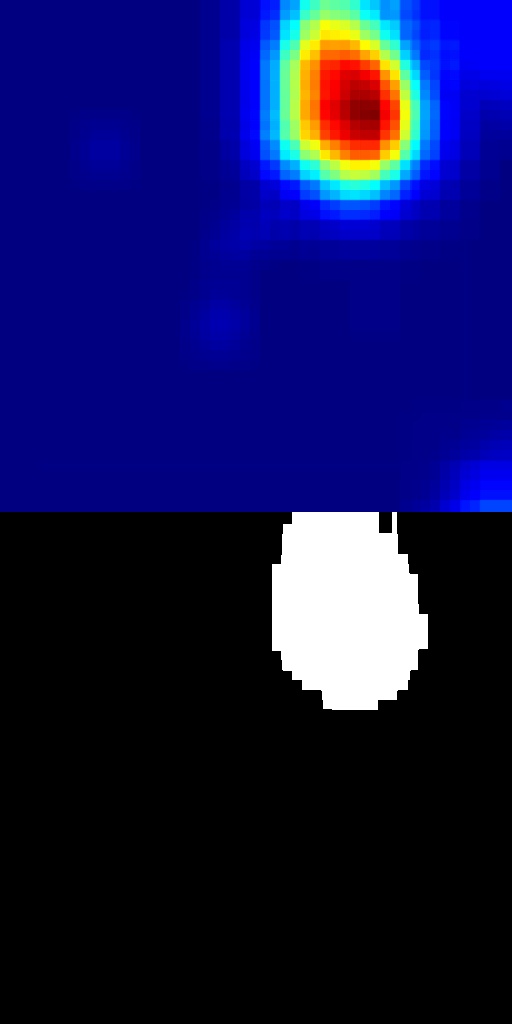}} \hfill
\subfloat{\includegraphics[width=0.195\linewidth]{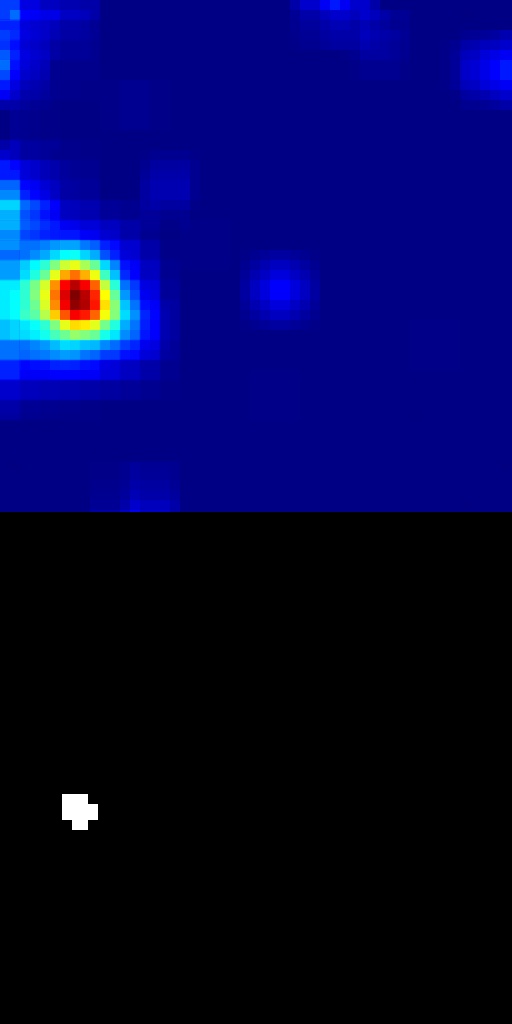}} \hfill
\subfloat{\includegraphics[width=0.195\linewidth]{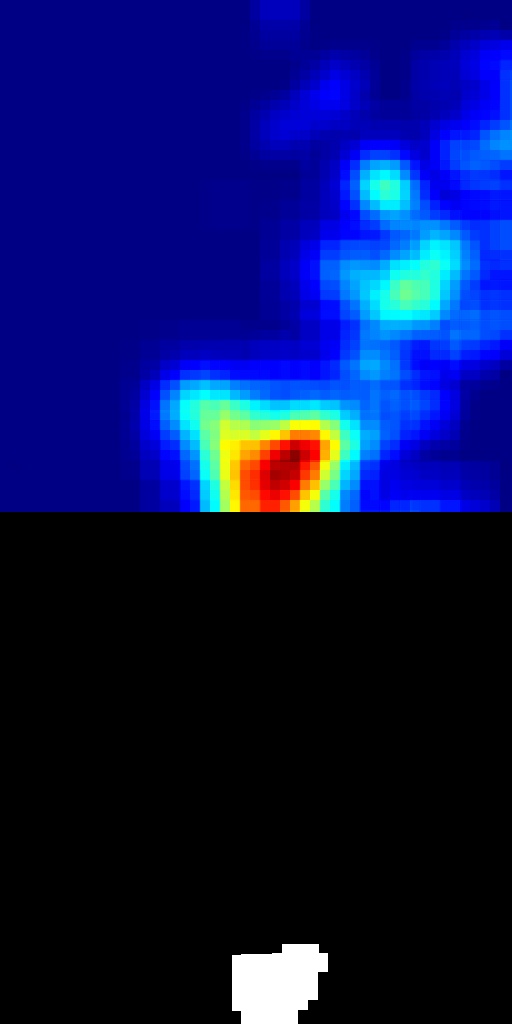}} \hfill
\subfloat{\includegraphics[width=0.195\linewidth]{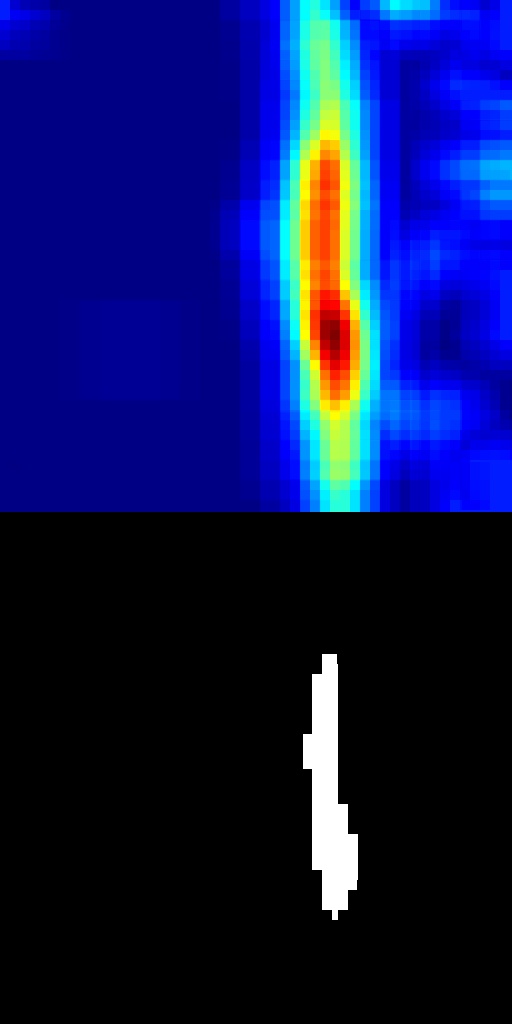}} \hfill
\subfloat{\includegraphics[width=0.195\linewidth]{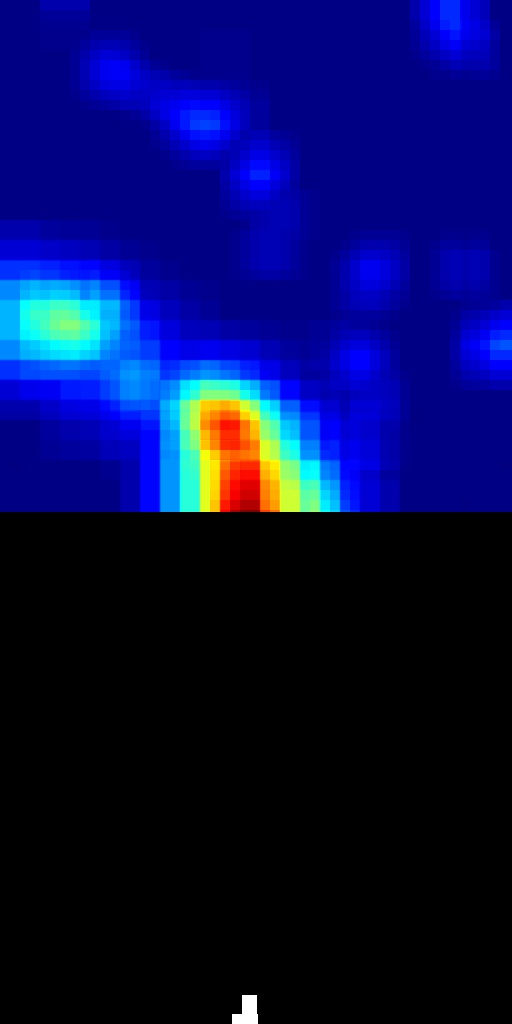}} \hfill \\[-20px]

\subfloat{\includegraphics[width=0.195\linewidth]{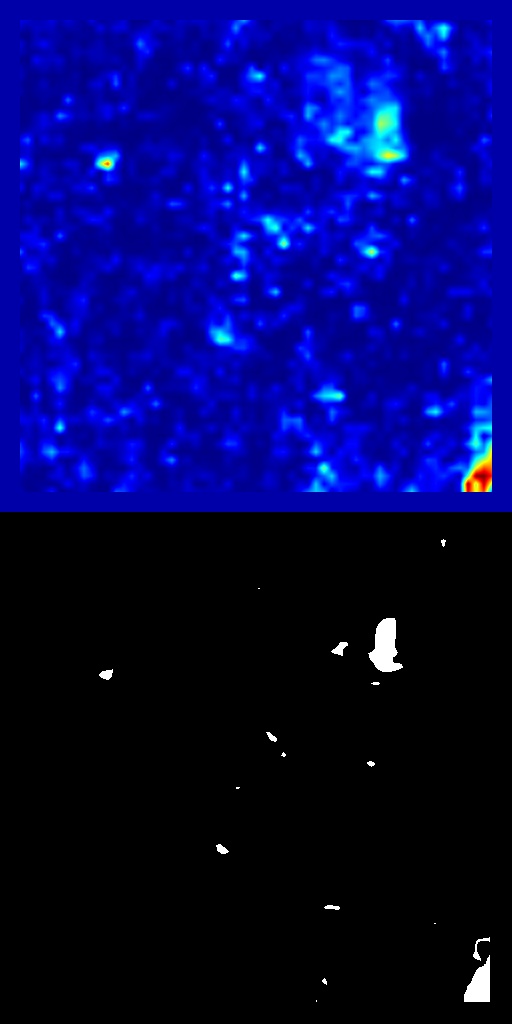}} \hfill
\subfloat{\includegraphics[width=0.195\linewidth]{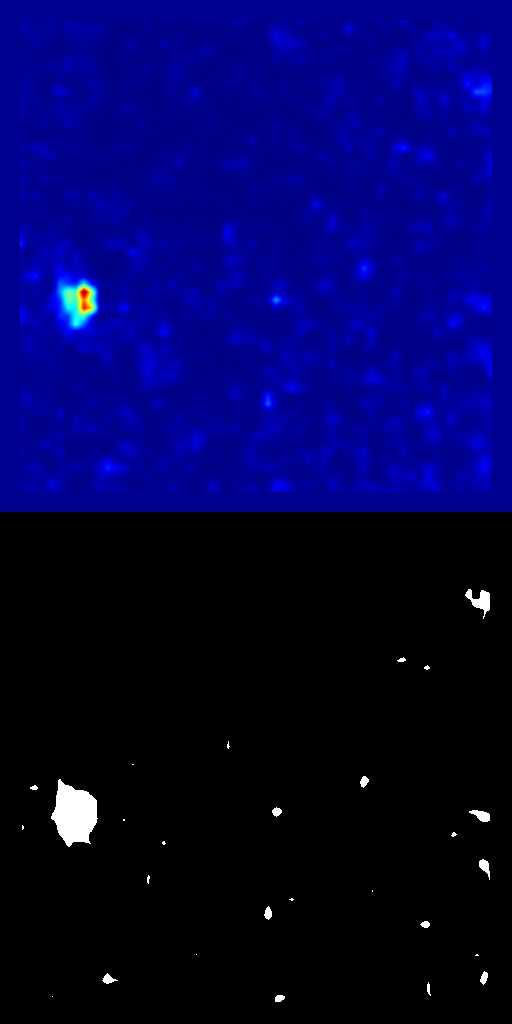}} \hfill
\subfloat{\includegraphics[width=0.195\linewidth]{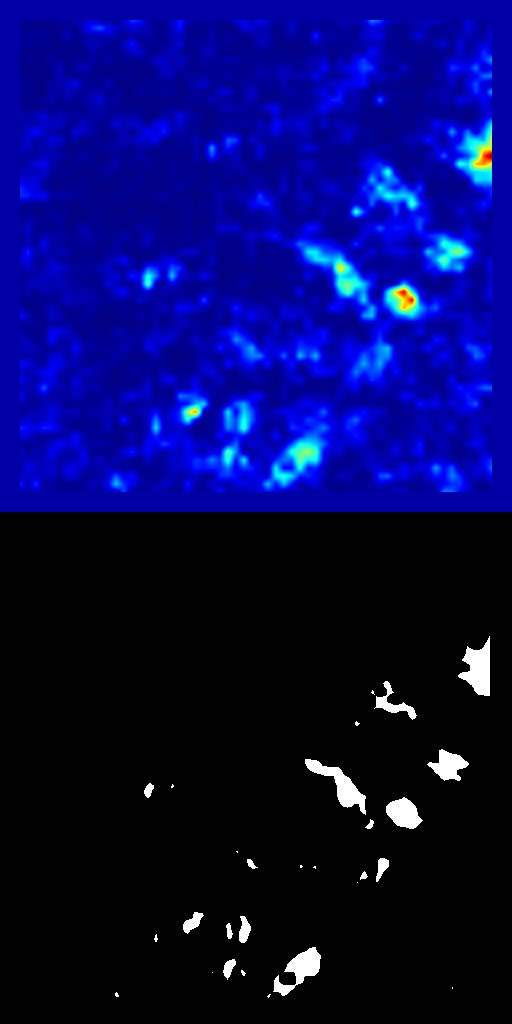}} \hfill
\subfloat{\includegraphics[width=0.195\linewidth]{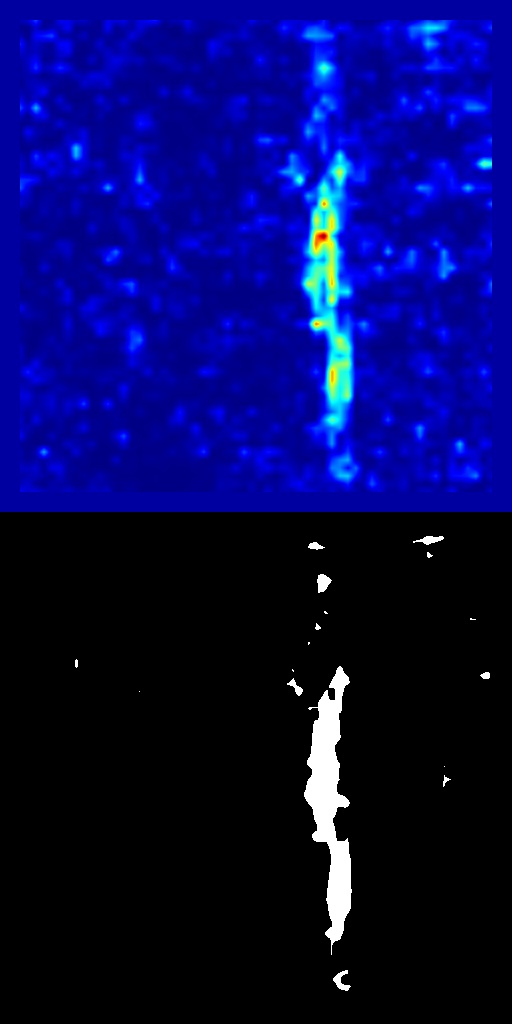}} \hfill
\subfloat{\includegraphics[width=0.195\linewidth]{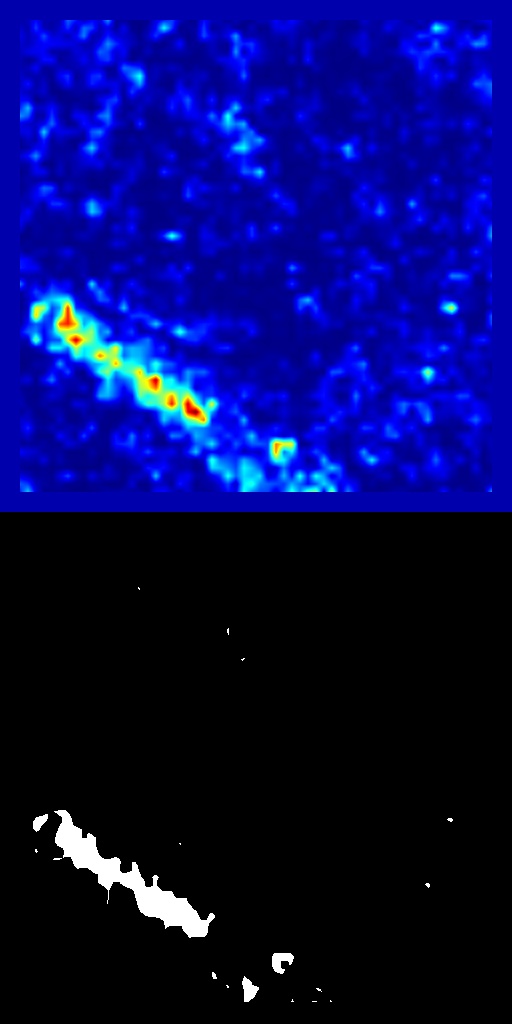}} \hfill \\[-20px]

\subfloat{\includegraphics[width=0.195\linewidth]{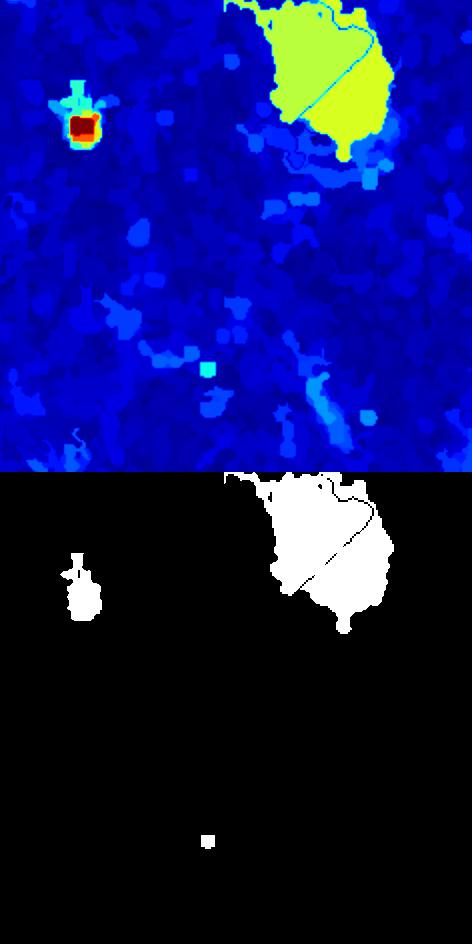}} \hfill
\subfloat{\includegraphics[width=0.195\linewidth]{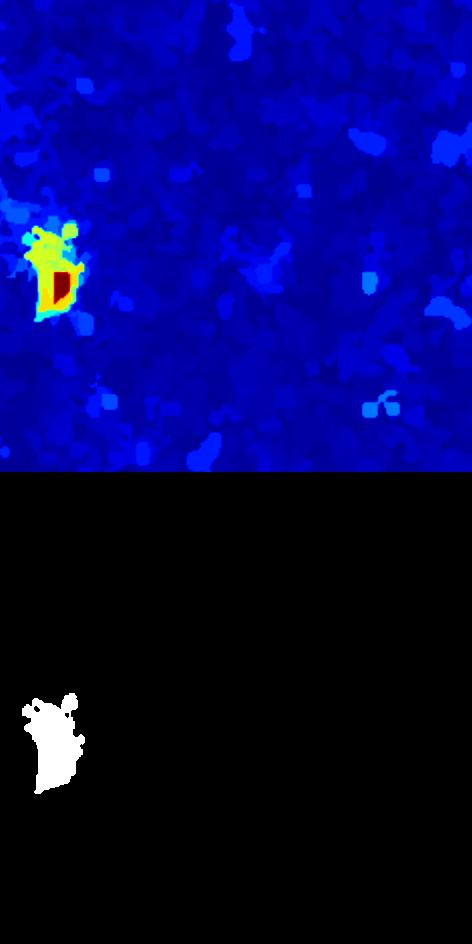}} \hfill
\subfloat{\includegraphics[width=0.195\linewidth]{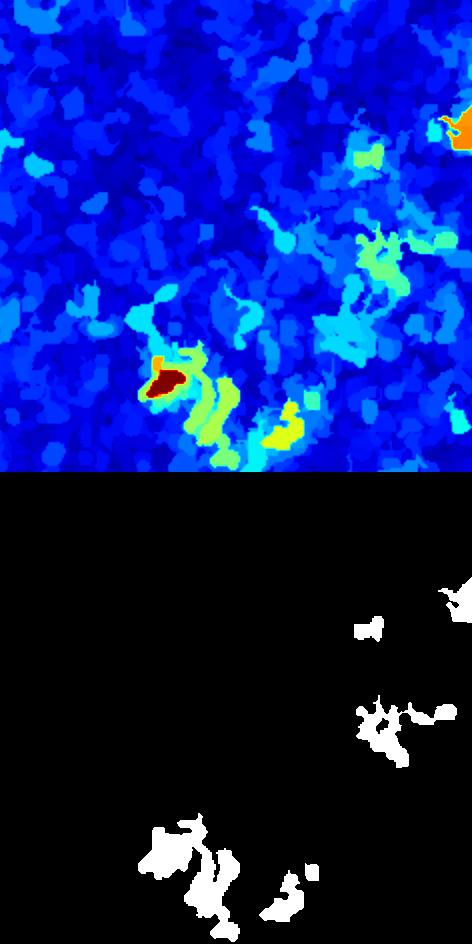}} \hfill
\subfloat{\includegraphics[width=0.195\linewidth]{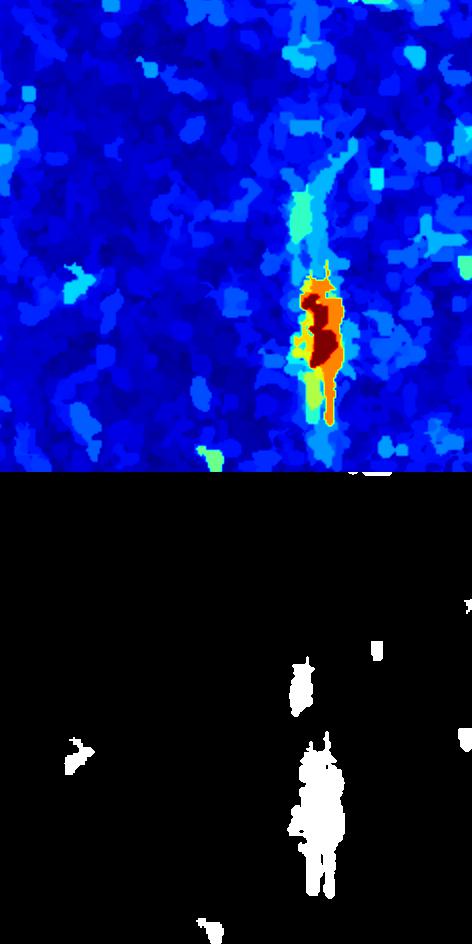}} \hfill
\subfloat{\includegraphics[width=0.195\linewidth]{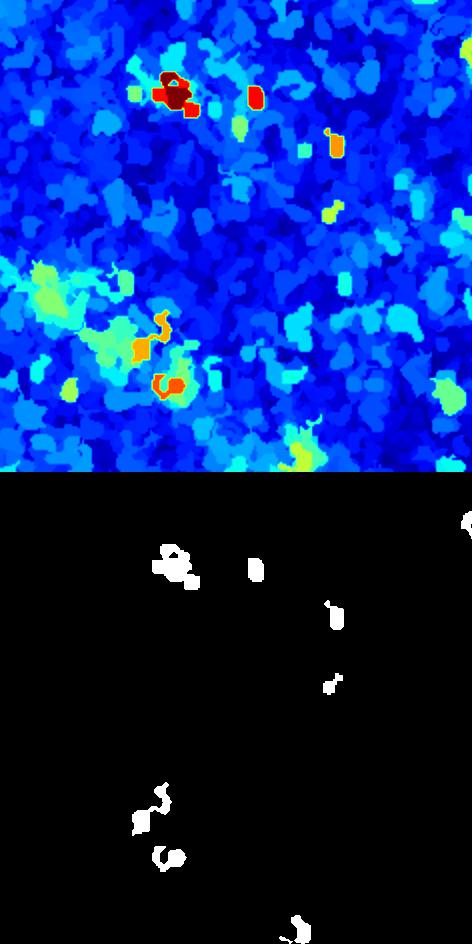}} \hfill \\%[-20px]

    \caption{Results on our industrial data. First row: original diffuse image. Following rows show the anomaly score map and the segmentation with AS=0, for some variants of our proposed method: \textit{PCA+PixelNFA}, \textit{Gabor+BlockNFA}, \textit{ResNet+PixelNFA}, and \textit{ResNet+RegionNFA}. All defects are detected in all cases, and AS=0 provides a good choice for the detection threshold.}
    %\vspace{-10pt}
    \label{fig:results-bader}
\end{figure}

%\textcolor{red}{Cambiamos el discurso sobre las anotaciones? Puede servir como argumento de otra cosa más que agregamos al otro paper?}
For this industrial application, no reliable anomaly segmentation is available. We only were able to gather a limited number of defective samples, that were coarsely annotated 
%and with notable errors by a operator,
during the normal inspection work. We present the ROC AUC results for all our method variants, and some examples for visual inspection in Figure~\ref{fig:results-bader}, showing the input image, its NFA maps and their corresponding segmentation with $\text{NFA}=0$ for some variants of the proposed method: \textit{PCA+PixelNFA}, \textit{Gabor+BlockNFA}, \textit{ResNet+PixelNFA} and \textit{ResNet+RegionNFA}. 
With all the variants of our method, we succeed in detecting all defects for this samples. For the Block-NFA variant we obtain a coarser map compared to Pixel-NFA, caused by the computation by blocks. On one hand, this characteristic makes the map cleaner and smoother, but in the other hand it may cause the evaluation metric used to drop, as it depends on a fine segmentation of the anomaly.

\begin{table}
\begin{center}
\normalsize
\caption{\vspace{2pt}ROC AUC results for the industrial application considered: defect detection in leather samples.}
\label{tab:resultsbader}
    \begin{tabular}{l l}
        \toprule
        Method  & ROC AUC  \\
        \midrule
        PCA+PixelNFA & 89.53\\
        PCA+BlockNFA & 78.63\\
        Gabor+PixelNFA & 85.67\\
        Gabor+BlockNFA & 78.41\\
        ResNet+PixelNFA & 87.84\\
        ResNet+RegionNFA & 87.92\\
        \bottomrule     
    \end{tabular}
\end{center}
\vspace{-12pt}
\end{table}

%\textcolor{blue}{
%\begin{itemize}
%    % \item PCA+PixelNFA: 83.03\%,
%    % \item PCA+BlockNFA: 84.20\%,
%    % \item Gabor+PixelNFA: 81.81\%,
%    % \item Gabor+BlockNFA: 85.78\%,
%    % \item ResNet+PixelNFA: 77.00\%,
%    % \item ResNet+RegionNFA: 80.39\%.
%    \item PCA+PixelNFA: 89.53\%,
%    \item PCA+BlockNFA: 78.63\%,
%    \item Gabor+PixelNFA: 85.67\%,
%    \item Gabor+BlockNFA: 78.41\%,
%    \item ResNet+PixelNFA: 87.84\%,
%    \item ResNet+RegionNFA: 87.92\%.
%\end{itemize}
%}

Since we have an input of 5 images (1 diffuse light and 4 grazing directional lights), we cannot directly compare other state of the art methods for this dataset, as they expect a single image as input. Moreover, for the ResNet variant of our method we had to include a modification, feeding the ResNet independently for each image and concatenating the output feature maps. Also, for these examples we use only the output of the first layer of ResNet, \textit{layer1/conv3}.
The ROC AUC results obtained are presented in Table~\ref{tab:resultsbader}. The best performance is achieved by the method PCA+PixelNFA, followed by ResNet+RegionNFA. The main drawback of PCA+PixelNFA is that we must compute the PCA for each image before filtering. On the other hand, ResNet+RegionNFA filtering is faster overall but we still need to compute the PCA after filtering. Gabor+PixelNFA, although its performance (in terms of AUC) is not so good, is a very simple method that does not require PCA after filtering.

We tested our algorithm in production with thousands of leather samples in the modality of assessment for the operator. More important than the results commented above, the first results obtained in production demonstrate that this tool provides useful assistance to the the operators, indicating problematic zones and making their work easier and faster.

% ------------------------------------------------------------------
\section*{Conclusions}
% ------------------------------------------------------------------

In this work we presented a fully unsupervised {\em a contrario} method for anomaly detection, which aims at detecting defects in leather samples from the automotive industry, achieving industry standards. Although the method was designed for this particular application, its foundations are general enough and proved to perform successfully in a wide variety of defects in images of different scenarios. The proposed approach outperforms other state-of-the-art methods on MVTec dataset, and has proven to work specially well for leather samples. From all the proposed variants, the one that clearly outstands is using the ResNet feature extraction with the Region NFA computation. It achieves the best score in almost all datasets, and it is ranked as the best overall. In addition, this method obtained the best separation between normal and defective samples, as can be deduced by the GAP analysis. Last but not least, the natural threshold (log-NFA=0) represents a good choice for easily fixing the detection threshold, as shown in Figure~\ref{fig:roc}.

Future potential improvements include substituting PCA by non linear transformations obtained from auto-encoders and/or normalizing flows trained on normal data. In this setting, the multi-scale approach may be included inherently in the network by means of a U-shaped network. Another line of work to be explored is the combination of the {\em a contrario} approach with deep neural networks in order to output detections with a confidence score, or to train the ResNet we utilized with leather anomaly-free samples, in order to obtain a more specific feature extraction for our task.

\captionsetup{position=top}
% \lipsum[1-2]
% \lipsum[3-10]
% \captionsetup{position=bottom}

% \input{plot_ipol}

% \begin{thebibliography}{00}
\bibliographystyle{IEEEtran}
\bibliography{nfa_icmla}
% \end{thebibliography}

\end{document}